%% file: main.tex
\documentclass{article}

\PassOptionsToPackage{numbers, compress}{natbib}

\usepackage[final]{neurips_2021}

\usepackage[utf8]{inputenc} 
\usepackage[T1]{fontenc}    
\usepackage{url}            
\usepackage{booktabs}       
\usepackage{amsfonts}       
\usepackage{nicefrac}       
\usepackage{microtype}      

\usepackage[usenames,dvipsnames,svgnames]{xcolor}
\usepackage{color}
\usepackage[colorlinks=true, linkcolor=red, urlcolor=NavyBlue, citecolor=NavyBlue]{hyperref}

\usepackage{amsmath}
\usepackage{algorithm}
\usepackage{algorithmic}
\usepackage{amssymb}
\usepackage{amsthm}
\usepackage{graphicx}
\usepackage{subfigure}
\usepackage{adjustbox}
\usepackage{hhline}
\usepackage{makecell}
\usepackage{enumitem}
\usepackage{soul}           
\usepackage{changepage}     

\setcitestyle{square}

\newcommand{\E}{\mathbb{E}}

\newcommand{\F}{\mathbb{F}}
\newcommand{\Q}{\mathbb{Q}}
\newcommand{\hF}{\hat{\mathbb{F}}}
\newcommand{\hQ}{\hat{\mathbb{Q}}}
\newcommand{\hp}{\hat{p}}

\newcommand{\pobs}{\hat{p}_{\textit{avg}}^{\text{obs}}}
\newcommand{\ppobs}{p_{\textit{avg}}^{\text{obs}}}

\newcommand{\cx}{\mathcal{X}}
\newcommand{\cy}{\mathcal{Y}}

\newcommand{\bX}{\textbf{X}}

\newcommand{\bY}{\textbf{Y}}

\newcommand{\eps}{\epsilon}
\newcommand{\sth}{\textsuperscript{th}}

\newcommand{\marginal}{\mathbb{F}_{\textbf{Y}}}

\newcommand{\toptitlebar}{
  \hrule height 4pt
  \vskip 0.25in
  \vskip -\parskip%
}
\newcommand{\bottomtitlebar}{
  \vskip 0.29in
  \vskip -\parskip
  \hrule height 1pt
  \vskip 0.09in%
}
\newcommand\blfootnote[1]{%
  \begingroup
  \renewcommand\thefootnote{}\footnote{#1}%
  \addtocounter{footnote}{-1}%
  \endgroup
}

\title{Beyond Pinball Loss: Quantile Methods for Calibrated Uncertainty Quantification}

\author{
    Youngseog Chung\\
    Machine Learning Department\\
    Carnegie Mellon University\\
    Pittsburgh, PA 15213\\
    \texttt{youngsec@cs.cmu.edu}
    \And
    Willie Neiswanger\\
    Department of Computer Science\\
    Stanford University\\
    Stanford, CA 94305\\
    \texttt{neiswanger@cs.stanford.edu}
    \And
    Ian Char\\
    Machine Learning Department\\
    Carnegie Mellon University\\
    Pittsburgh, PA 15213\\
    \texttt{ichar@cs.cmu.edu}
    \And
    Jeff Schneider\\
    Robotics Institute\\
    Carnegie Mellon University\\
    Pittsburgh, PA 15213\\
    \texttt{schneide@cs.cmu.edu}
    
}

\begin{document}

\maketitle

\input{abstract}

\input{introduction}

\input{prelim}

\input{methods}

\input{experiments}

\input{conclusion}
\input{acknowledgments}

\newpage
\bibliographystyle{plainnat}
\bibliography{refs}



\input{appendix}

\end{document}

%% file: abstract.tex
\vspace{-3mm}
\begin{abstract}
\vspace{-1mm}

    Among the many ways of quantifying uncertainty in a regression setting, 
    specifying the full quantile function is attractive, 
    as quantiles are amenable to interpretation and evaluation. 
    A model that predicts the true conditional quantiles for each input, 
    at all quantile levels, presents a correct and efficient representation of the underlying uncertainty.
    To achieve this, many current quantile-based methods focus on optimizing the
    \textit{pinball loss}. However, this loss restricts the scope of applicable regression models,
    limits the ability to target many desirable properties (e.g. calibration, sharpness, centered intervals), 
    and may produce poor conditional quantiles.
    In this work, we develop new quantile methods that address these shortcomings.
    In particular, we propose methods that can apply to any class of regression model, 
    select an explicit balance between calibration and sharpness, 
    optimize for calibration of centered intervals, and produce more accurate conditional quantiles.
    We provide a thorough experimental evaluation of our methods,
    which includes a high dimensional uncertainty quantification task in nuclear fusion.
    \blfootnote{Code is available at \url{https://github.com/YoungseogChung/calibrated-quantile-uq}.}

\end{abstract}


%% file: introduction.tex
\vspace{-3mm}
\section{Introduction} 
\label{sec:introduction}
\vspace{-1mm}

Uncertainty quantification (UQ) in
machine learning typically refers to the task of quantifying the confidence of a given prediction.
This measure of certainty can be crucial in a variety of downstream applications, including
Bayesian optimization \citep{kushner1964new, mockus1978application, shahriari2015taking},
model-based reinforcement learning \citep{yu2020mopo, malik2019calibrated, chua2018deep, garcia2012safe},
and in high-stakes predictions where mistakes incur large costs
\citep{rudin2019stop, wexler2017computer}.

While the common goal of UQ is to describe predictive distributions over outputs for given inputs,
the representation of the distributional prediction varies across methods.
For example, some methods assume a parametric distribution and return parameter estimates \citep{zhao2020individual, detlefsen2019reliable, lakshminarayanan2017simple}, 
while others return density function estimates, as is common in Bayesian methods 
\citep{maddox2019simple, liu2019accurate, hernandez2015probabilistic, blundell2015weight, koller2009probabilistic, rasmussen2003gaussian}. 
Alternatively, many methods represent predictive uncertainty with quantile estimates \citep{fasiolo2020fast, salem2020prediction, tagasovska2019single, pearce2018high, jeon2016quantile}.

Quantiles provide an attractive representation for uncertainty because they
can be used to model complex distributions without parametric assumptions,
are interpretable with units in the target output space,
allow for easy construction of prediction intervals, 
and can be used to efficiently sample from the predictive distribution via
inverse transform sampling \citep{hao2007quantile, koenker2001quantile}.
Learning the quantile for a single quantile level is a well studied problem
in quantile regression (QR) \citep{koenker2005quantile, koenker2001quantile},
which typically involves optimizing the so-called \textit{pinball loss}, 
a tilted transformation of the absolute value function. Given a target $y$, a prediction $\hat{y}$, and quantile level $\tau \in (0,1)$, the pinball loss $\rho_{\tau}$ is defined as
\begin{align} \label{eq:pinball_loss}
    \rho_{\tau}(y, \hat{y}) = (\hat{y} - y)(\mathbb{I}\{y \leq \hat{y}\} - \tau).
\end{align} 
By training for all quantiles simultaneously, recent works have made concrete steps in incorporating QR methods 
to form competitive UQ methods which output the full predictive distribution \citep{rodrigues2020beyond, tagasovska2019single}.


In this work, we highlight some limitations of the pinball loss and propose several methods to address these shortcomings.
Specifically, we explore the following:
\begin{itemize}[topsep=0mm, leftmargin=5mm]
    \item \textbf{Model agnostic QR.} 
    Optimizing the pinball loss often restricts the choice of model family for which we can
    provide UQ.
    We propose an algorithm to learn all quantiles simultaneously by utilizing methods from 
    conditional density estimation. 
    This algorithm is agnostic to model class and can be applied to \textit{any} regression model.
    \item \textbf{Explicitly balancing calibration and sharpness.} 
    While the pinball loss, as a proper scoring rule, targets both calibration and sharpness, the balance between these two quantities is made implicitly, which may 
    result in a poor optimization objective.
    We propose a tunable loss function that targets calibration and sharpness separately,
    and allows the end-user to set an \textit{explicit} balance.
    \item \textbf{Centered intervals.} 
    In practice, we often desire uncertainty predictions made with centered intervals, which
    are not targeted via the pinball loss.
    We propose an alternative loss function that is better suited for this goal.
    \item \textbf{Encouraging individual calibration.}
    Perfect quantile forecasts will satisfy \textit{individual} calibration (Eq.~\ref{eq:indv_cali}), 
    which is a much stricter condition than the
    more-commonly used notion of \textit{average} calibration (Eq.~\ref{eq:avg_cali}).
    We introduce a training procedure that aims to improve quantile predictions
    beyond average calibration, and demonstrate its efficacy via \textit{adversarial group}
    calibration (Eq.~\ref{eq:adversarial-group-calibration}).
\end{itemize}
We proceed by first describing methods of assessing the quality of predictive UQ 
and the pitfalls of optimizing the pinball loss in Section~\ref{sec:prelim}.
Drawing motivation from this, we then present our proposed methods in Section~\ref{sec:methods}.
In Section~\ref{sec:experiments}, we demonstrate our methods experimentally,
where we model predictive uncertainty on benchmark datasets, and 
on a high-dimensional, real-world uncertainty estimation task in the area of nuclear fusion.

%% file: prelim.tex
\vspace{-2mm}
\section{Preliminaries and Background}
\label{sec:prelim}
\vspace{-2mm}

We first lay out the notation, terminology, and class of models considered in this paper.
Then we provide an overview of evaluation metrics in UQ and demonstrate 
how the pinball loss may be inadequate both as an evaluation metric 
and as an optimization objective.

\vspace{-1mm}
\subsection{Notation} \label{sec:notation}
\vspace{-2mm}
Bold upper case letters \textbf{X, Y} denote random variables, 
lower case letters $x, y$, denote their values, 
and calligraphic upper case letters $\mathcal{X, Y}$ denote sets of possible values. 
We use $x\in \mathcal{X}$ to denote the input feature vector and $y \in \mathcal{Y}$ to denote the corresponding target.
Additionally, we consider the regression setting where $\cy \subset \mathbb{R}$ and $\mathcal{X} \subset \mathbb{R}^n$.
We use $\mathbb{F}_{\textbf{X}}, \mathbb{F}_{\textbf{Y}\mid x}, \mathbb{F}_{\textbf{Y}}$ 
to denote the true cumulative distribution of the subscript random variable. 
For any $x \in \mathcal{X}$, we assume there exists a true conditional distribution 
$\mathbb{F}_{\textbf{Y}\mid x}$ over $\cy$, 
and we assume $\Q_p(x)$ denotes the true $p$\textsuperscript{th} quantile of this distribution, 
i.e. $\mathbb{F}_{\textbf{Y}\mid x}(\Q_p(x)) = p$.
Any estimates of the true functions $\mathbb{F}, \Q_p$ will be denoted with a hat, 
$\hat{\mathbb{F}}, \hQ_p$.
We will specifically refer to any family of estimates for $\Q_p$, with $p \in (0, 1)$,
as a ``quantile model'', denoted
$\hQ: \mathcal{X} \times (0, 1) \rightarrow \mathcal{Y}$.
Unless otherwise noted, we will always consider the \textit{conditional} problem of
estimating quantities in the target space $\cy$, conditioned on a value $x \in \cx$.


\vspace{-1mm}
\subsection{Assessing the Quality of Predictive UQ}
\label{sec:metrics}
\vspace{-2mm}







While various metrics have been
proposed to assess the quality of UQ, there has been a great deal of recent focus on the notions of \textit{calibration} and \textit{sharpness}
\citep{fasiolo2020fast, cui2020calibrated, zhao2020individual, tran2020methods, song2019distribution, kuleshov2018accurate, guo2017calibration, gneiting2007probabilistic}.
We introduce calibration here, but for a more thorough treatment, see
\citet{zhao2020individual}.
Broadly speaking, calibration in the regression setting
requires that the probability of observing the target random variable below a predicted $p$\textsuperscript{th} quantile is equal to the \textit{expected probability} $p$, for all $p \in (0, 1)$.
We refer to the former quantity as the \textit{observed probability} and denote it 
$p^{\text{obs}}(p)$, for an expected probability $p$, which we will write as
$p^{\text{obs}}$ when it is clear from context.
Calibration requires $p^{\text{obs}}(p) = p$, $\forall p \in (0,1)$.
From this generic statement, we can describe different notions of calibration
based on how $p^{\text{obs}}$ is defined.

A model is \textbf{individually calibrated} if it outputs the true conditional quantiles, i.e. $\hQ_p(x) = \Q_p(x)$. In this case, 
we define the observed probability to be
\begin{align} \label{eq:indv_cali}
    p^{\text{obs}}_{indv}(p, x) := \mathbb{F}_{\textbf{Y}|x}(\hQ_p(x)),
    \hspace{2mm} \forall x \in \cx, \hspace{2mm} \forall p \in (0, 1).    
\end{align} 
In words, this requires that the probability of observing $y$ below the quantile prediction is equal to $p$, \textit{at each point x $\in \cx$, individually}.
If we can verify this property for all $x \in \cx$, then by definition, 
we will know the quantile output is correct and precisely the true conditional quantile. 
However, individual calibration is typically unverifiable with finite datasets in the assumption-less case \citep{zhao2020individual}. 

A relaxed condition is \textbf{average calibration}.
In this case, we define the observed probability to be
\begin{align} \label{eq:avg_cali}
    p^{\text{obs}}_{avg}(p) := \mathbb{E}_{x \sim \F_{\textbf{X}}}[\mathbb{F}_{\textbf{Y}|x}(\hQ_p(x))], \hspace{2mm} \forall p\in (0, 1),
\end{align}
i.e. the probability of observing the target below the quantile prediction, 
\textit{averaged over $\F_{\textbf{X}}$}, is equal to $p$. 
Average calibration is often referred to simply as ``calibration''
\citep{cui2020calibrated, kuleshov2018accurate}.
Given a dataset $D = \{(x_i, y_i)\}_{i=1}^{N}$, we can estimate $p^{\text{obs}}_{avg}(p)$ with
$\hat{p}^{\text{obs}}_{avg}(D, p) =
\frac{1}{N}\sum_{i=1}^{N} \mathbb{I}\{y_i \leq \hQ_p(x_i)\}$.
%
%
Note that if our quantile estimate achieves average calibration then $\hat{p}^{\text{obs}}_{avg} \rightarrow p$ as $N\rightarrow\infty$, $\forall p \in (0, 1)$.
%
%
The degree of error in average calibration is commonly measured by \textit{expected calibration error} \cite{tran2020methods, cui2020calibrated, guo2017calibration},
$\text{ECE}(D, \hQ) = \frac{1}{m}\sum_{j=1}^{m} \left | \hat{p}^{\text{obs}}_{avg}\left(D, p_j\right) - p_j \right |$,
where $p_j \sim \text{Unif}(0, 1)$. 

It may be possible to have an uninformative, yet average calibrated model. For example,
quantile predictions that match the true \textit{marginal} quantiles of $\F_{\textbf{Y}}$ 
will be average calibrated, but will hardly be useful since they do not depend on the input $x$.
Therefore, the notion of \textbf{sharpness} is also considered, which quantifies the concentration of distributional predictions \citep{gneiting2007probabilistic}.
For example, for non-parametric predictions, the width of a centered $95\%$ prediction interval is often used as a measure of sharpness.
There generally exists a tradeoff between average calibration and sharpness \citep{gneiting2007probabilistic, murphy1973new}.

Recent works have suggested a notion of calibration stronger than average calibration, called adversarial group calibration \citep{zhao2020individual}.
This stems from the notion of \textbf{group calibration} \citep{hebert2017calibration, kleinberg2016inherent}, which prescribes measurable subsets 
$\mathcal{S}_i \subset \cx$ s.t. $P_{x \sim \F_{\textbf{X}}}(x \in \mathcal{S}_i) > 0$, $i=1,\dots,k$, 
and requires the predictions to be average calibrated within each subset.
Adversarial group calibration then
requires average calibration for \textit{any subset of $\cx$ with non-zero measure}.
Denote $\textbf{X}_{\mathcal{S}}$ as a random variable that is conditioned on being in the set $\mathcal{S}$.
For \textbf{adversarial group calibration}, the observed probability is
\begin{align} \label{eq:adversarial-group-calibration}
\begin{split}
    %
    %
    p^{\text{obs}}_{adv}(p) := \hspace{2mm} \mathbb{E}_{x \sim \F_{\textbf{X}_{\mathcal{S}}}}[\mathbb{F}_{\textbf{Y}|x}(\hQ_p(x))],
    \hspace{2mm} \forall p \in (0, 1),
    \hspace{2mm} \forall \mathcal{S} \subset \cx \text{  s.t.  } P_{x \sim \F_{\textbf{X}}} (x \in \mathcal{S}) > 0.
\end{split}
\end{align}
With a finite dataset, we can measure a proxy of adversarial group calibration by measuring the average calibration within all subsets of the dataset with sufficiently many points.

Intuitively, individual calibration inspects the discrepancy between $p^{\text{obs}}$ and $p$ for
individual inputs $x \in \cx$, adversarial group calibration relaxes this by
inspecting any subset of $\mathcal{X}$ with non-zero measure, and
average calibration relaxes this further by considering the full distribution of $\bX$.


\begin{figure}[t!]
    \centering
    \subfigure[Test Loss Curves]{\includegraphics[width=0.325\textwidth]{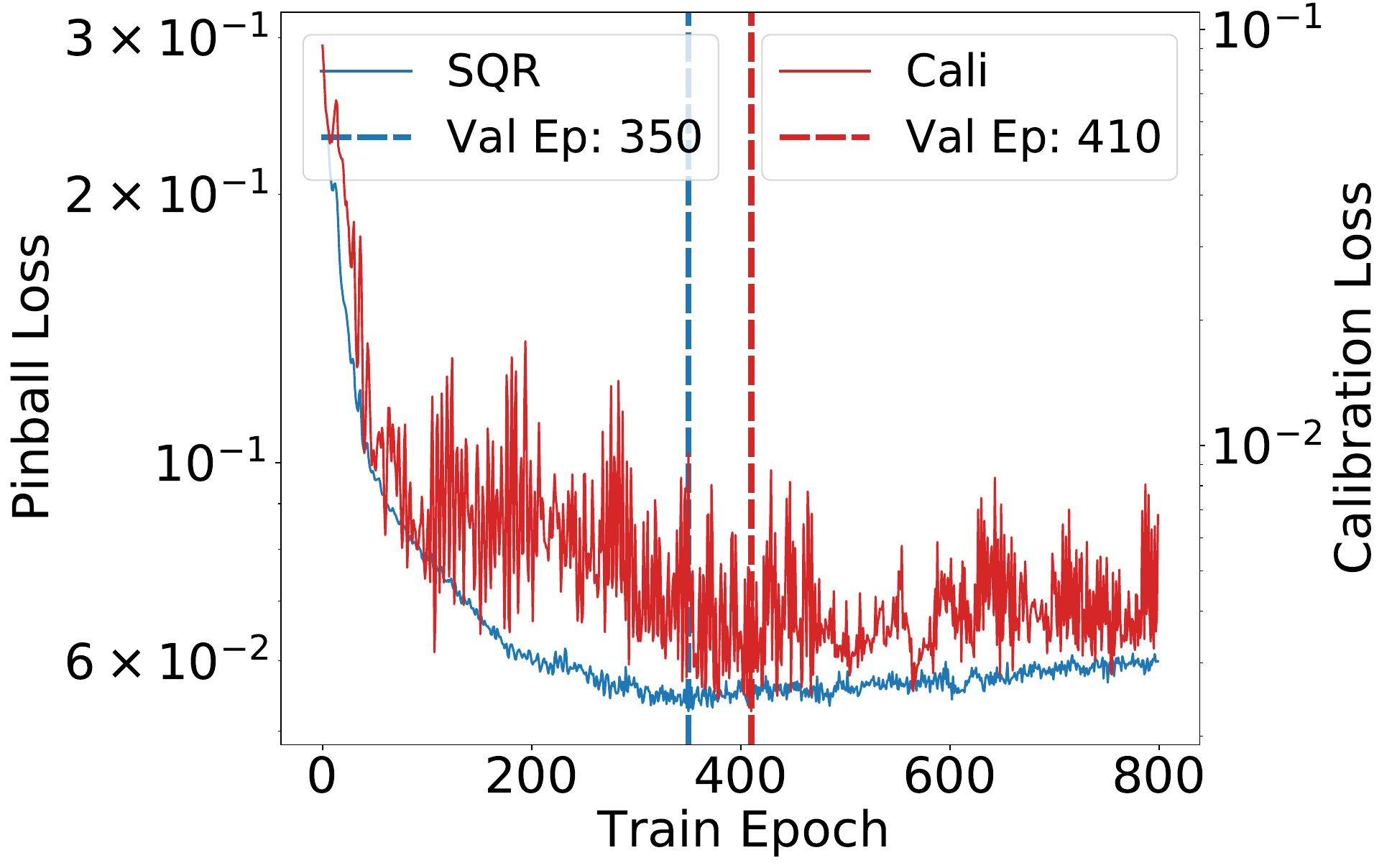}}
    \subfigure[Test Calibration]{\includegraphics[width=0.33\textwidth]{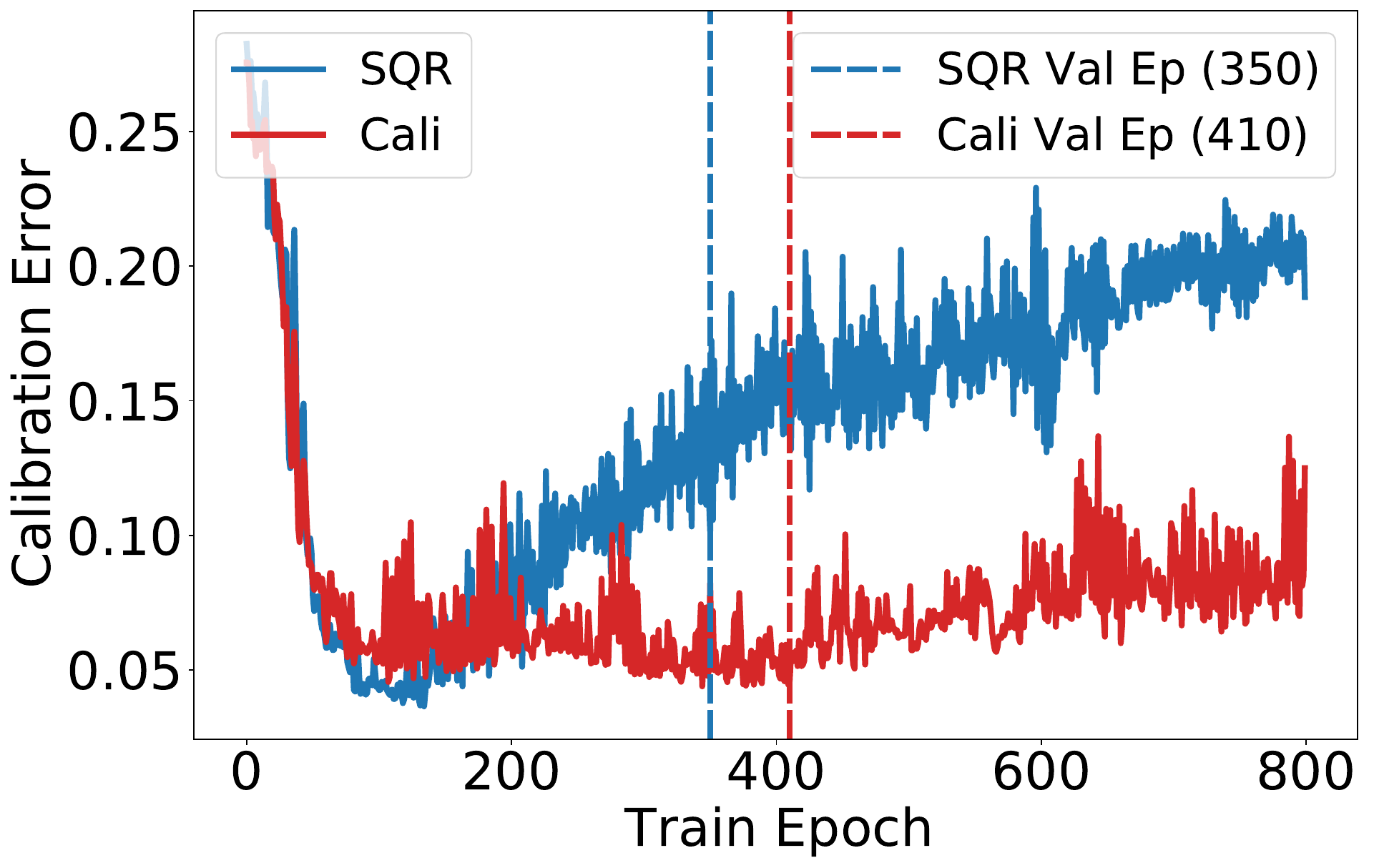}}
    \subfigure[Test Sharpness]{\includegraphics[width=0.33\textwidth]{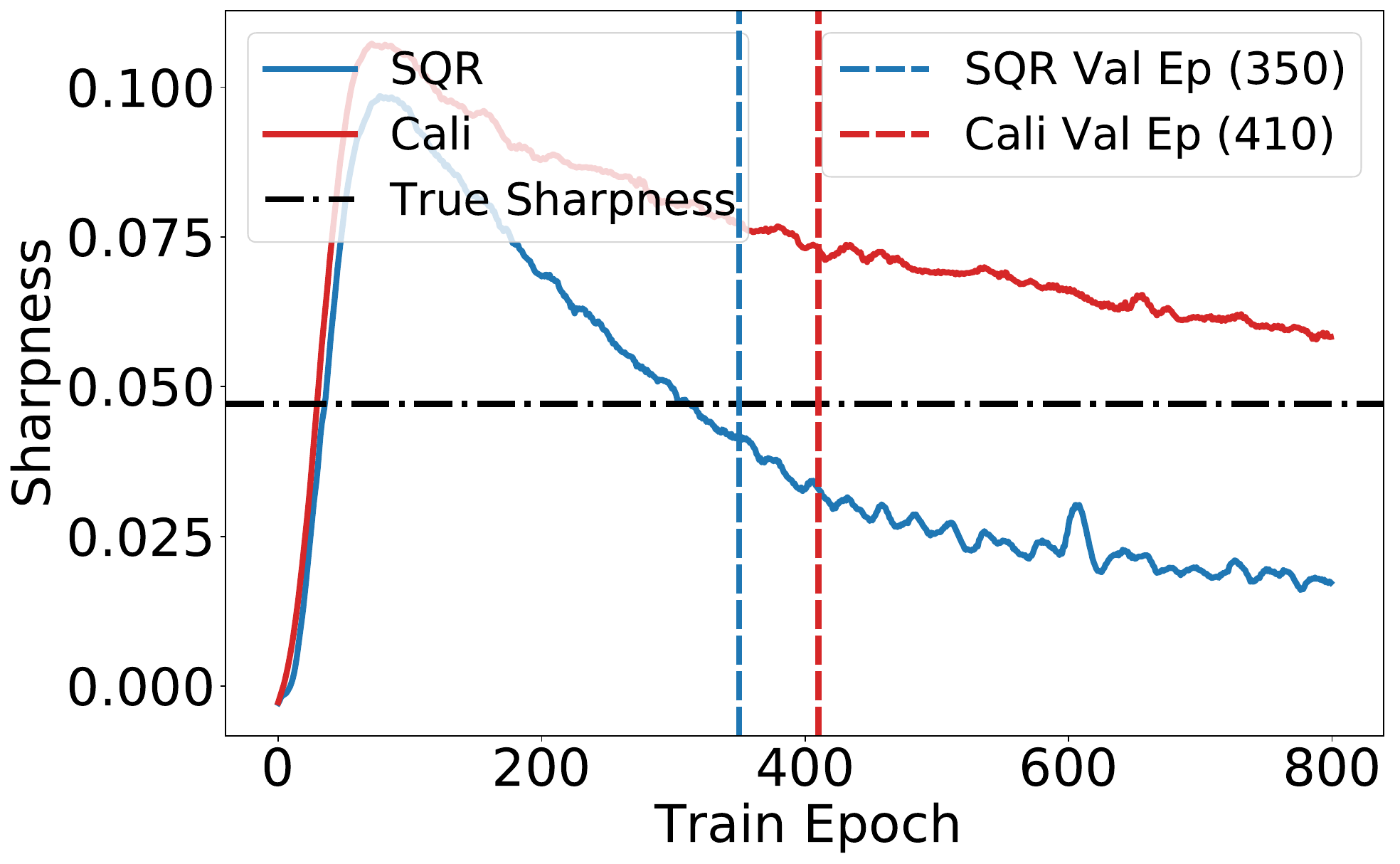}}
    \vspace{-3mm}
    \caption{\textbf{(a)} Test loss continues to decrease until the validated epoch. \textbf{(b-c)} At the validated epoch, \textit{SQR} (optimizes \textit{pinball loss}) is highly miscalibrated while sharper than the true sharpness level. \textit{Cali} (optimizes proposed \textit{calibration loss}) is better calibrated while less sharp than the true sharpness.}
    \label{fig:synth_example}
    \vspace{-2mm}
\end{figure}

One alternative family of evaluation metrics is \textbf{proper scoring rules} \citep{gneiting2007strictly}. 
Proper scoring rules are summary statistics of overall performance of a
distributional prediction and consider both calibration and sharpness
jointly \citep{gneiting2007probabilistic}. 
For example, negative log-likelihood (NLL) is a proper scoring rule
that is commonly used with density predictions \citep{detlefsen2019reliable, pearce2018uncertainty, lakshminarayanan2017simple}. 
For quantile predictions, one proper score is the \textbf{check score}, \textit{which is identical to the pinball loss}. 
Since proper scoring rules consider both calibration and sharpness together 
in a single value, they can serve as optimization objectives for UQ. 
For example, optimizing the pinball loss 
is the traditional method in quantile regression \citep{koenker1978regression},
and many recent quantile-based UQ methods focus on optimizing this objective \citep{rodrigues2020beyond, tagasovska2019single, cannon2018non, xu2017composite}.


In this work, however, we note that the balance between calibration and
sharpness implied by the pinball loss
is arbitrary and depends on the expressivity of the model class---and with highly
expressive models, this balance can be heavily skewed towards sharpness.
In their seminal work on probabilistic forecasts, 
\citet{gneiting2007strictly} contend that the goal of probabilistic forecasting 
is to ``maximize the sharpness of the predictive distribution subject to calibration'',
i.e. calibration should be first achieved and then sharpness optimized.
We show that common machine learning methods that use 
the pinball loss objective may in fact lead to an arbitrary and miscalibrated UQ.
%



{\bf Proposition 1}. \label{prop:pinball}
{\it Consider a finite dataset $D$, the pinball loss $\rho_\tau$ (Eq.~\ref{eq:pinball_loss}) and a quantile model
$f: \mathcal{X} \times (0, 1) \rightarrow \mathcal{Y}$ that is average calibrated on $D$, i.e. $\text{ECE}(D, f) = 0$.
Then there always exists another quantile model $g: \mathcal{X} \times (0, 1) \rightarrow \mathcal{Y}$, such that, for any quantile level $\tau \in (0, 1)$, 
$g$ has lower pinball loss than $f$ on $D$, i.e. 
$\sum_{i=1}^{N}\rho_\tau(y_i, g_\tau(x_i)) < \sum_{i=1}^{N}\rho_\tau(y_i, f_\tau(x_i))$,
but worse average calibration than $f$, i.e. $\text{ECE}(D, g) > \text{ECE}(D, f)$.} 

\textbf{Proof:} The proof is given in Appendix~\ref{app:prop_1}.




This proposition essentially states
how the pinball loss can become detached from
calibration, and we show its practical ramifications via a synthetic example in
Figure~\ref{fig:synth_example} (experiment details in Appendix~\ref{app:synth_example}).
We first note in Figure~\ref{fig:synth_example} (a) and (b) that
even while the pinball loss decreases on the test set,
test calibration worsens (while sharpness improves).
Further, at the best validation epoch, 
optimizing the pinball loss converges to a solution that is sharper than the true noise level. 
Note that a UQ that is sharper than the true noise level will \textit{never} be calibrated 
(meanwhile, a less sharp prediction \textit{can still be calibrated}, e.g. 
the marginal distribution $\F_{\textbf{Y}}$).
While this may seem like an issue that can simply be addressed with regularization, we demonstrate in Appendix~\ref{app:regularization_pinball} how that is not the case.
These pitfalls motivate our methods in Section~\ref{sec:methods}. 


%% file: methods.tex
\vspace{-1mm}
\section{Methods}
\label{sec:methods}
\vspace{-1mm}
We propose four methods that aim to produce an improved quantile model.
The first is a model-agnostic procedure that relies on conditional density estimation (Section~\ref{sec:maqr}).
To address settings where density estimation may be difficult, 
we then propose two loss functions to optimize with differentiable models: 
the combined calibration loss (Section~\ref{sec:cali}), which directly optimizes calibration and sharpness, and the interval score (Section~\ref{sec:int}), which is a proper scoring rule for centered intervals.
Finally, we propose a group batching method (Section~\ref{sec:group_batch}) that
can be applied to the batch optimization procedure for any loss function
(e.g. combined calibration loss, interval score, and even pinball loss) to induce better convergence towards adversarial group calibration.


\vspace{-1mm}
\subsection{Utilizing Conditional Density Estimation for Model Agnostic QR} \label{sec:maqr}
\vspace{-1mm}
One drawback of many existing quantile-based UQ methods is that 
their training procedure requires differentiable models. 
In fact, most UQ methods require a specific class of models because of
their modeling structure or their loss objective (e.g. Gaussian processes \citep{rasmussen2003gaussian}, dropout \citep{gal2016dropout}, latent variable models \citep{koller2009probabilistic}, simultaneous pinball loss \citep{tagasovska2019single},
and NLL-based losses \citep{lakshminarayanan2017simple}).
This model restriction can be especially unfavorable in practical settings. 
A domain expert with an established point prediction model and compute infrastructure may want to add UQ without much additional overhead.

To address these issues, we can consider the following model-agnostic procedure. 
Instead of optimizing a designated loss function, 
we can consider splitting the given problem into two parts: 
estimate conditional quantiles directly from data, 
then regress onto these estimates. 
The benefit of this method is that, granted we can estimate the
conditional quantiles accurately, 
we can use any regression model to regress onto these quantile estimates.
Further, this regression task directly targets the goal of producing
the true conditional quantiles (i.e. individual calibration).
This procedure, which we refer to as \textit{Model Agnostic QR} (MAQR), is outlined in Algorithm~\ref{alg:maqr_main}.

MAQR is based on the key assumption that 
nearby points in $\mathcal{X}$ will have similar conditional distributions, 
i.e. if $x_j \approx x_k$ then $\F_{\bY|x_j} \approx \F_{\bY|x_k}$. 
Given this smoothness assumption,
we can group neighboring points to estimate the conditional density at each locality over $\cx$,
with locality determined by the hyperparameter $d_N$ (Algorithm~\ref{alg:CondQuantEst}, line 2).
We then construct an empirical CDF with the group of neighboring points, 
and conditional quantile estimates are produced with this empirical CDF.
These estimates are collected into $D$ (Algorithm~\ref{alg:maqr_main}, line 6), 
which is ultimately used as the training set for the quantile model $\hat{g}$.

In practice, we perform these steps with \textit{residuals},
by first estimating a mean function $\hat{f}$ (Algorithm~\ref{alg:maqr_main}, line 1).
This practical choice stems from existing works in conditional density estimation, 
which suggests that having 0 conditional mean in the data provides benefits in terms of 
lower asymptotic mean squared error in the conditional density predictions~\citep{hyndman1996estimating}. 
Further, this demonstrates how MAQR can be readily applied in the application setting
where an accurate point prediction model often already exist.

\begin{figure*}[ttt!]
\begin{minipage}{0.48\textwidth}
\begin{algorithm}[H]
    \caption{\textsc{MAQR}} 
    \begin{algorithmic}[1] \label{alg:maqr_main}

        \STATE {\bfseries Input:} Train data $\{x_i, y_i\}_{i=1}^{N}$,
        trained regression model $\hat{f}(x)$
        
        \STATE Calculate residuals $\epsilon_i = y_i - \hat{f}(x_i)$, \hspace{1mm}$i\in[N]$, and denote the residual dataset $R = \{x_i, \eps_i\}_{i=1}^{N}$
        
        \STATE Initialize $D \leftarrow \varnothing$
        
        \FOR{$i=1$ {\bfseries to} $N$}
            \STATE $D_i$ $\leftarrow$ \textsc{CondQuantilesEstimators}($R$, $i$) (Algorithm 2) 
            \STATE $D \leftarrow D \cup D_i$
        \ENDFOR 
    
        \STATE Use $D$ to fit a regression model $\hat{g}$ \\
        \hspace{10pt} $\hat{g}: (x, p) \mapsto \eps$

        \STATE {\bfseries Output:} $\hat{f} + \hat{g}$
    \end{algorithmic}
\end{algorithm}
\end{minipage}
\hfill
\begin{minipage}{0.48\textwidth}
\begin{algorithm}[H]
    \caption{\textsc{CondQuantilesEstimators}} 
    \begin{algorithmic}[1] \label{alg:CondQuantEst}

        \STATE {\bfseries Input:} Dataset $\{x_i, \eps_i\}_{i=1}^{N}$, 
        point index $k\in[N]$
        
        \STATE $E_{k,{d_N}} \leftarrow \{\eps_i: \textrm{dist}(x_k, x_i) \leq d_N, i\in[N] \}$
    
        \STATE Construct an empirical CDF with $E_{k,d_N}$ to produce $\hF_{\mathbf{E}|x_k}: \epsilon \mapsto p\in[0,1]$
        
        \STATE Initialize $D \leftarrow \varnothing$
        \FOR{each $\eps_j$ in $E_{k, d_N}$}
            \STATE $\hat{p}_{k, j} \leftarrow \hF_{\mathbf{E}|x_k}(\eps_j)$
            \STATE $D \leftarrow D \cup \{x_k, \hat{p}_{k, j}, \eps_j\}$
        \ENDFOR
    \STATE {\bfseries Output:} $D$
    
    \end{algorithmic}
\end{algorithm}
\end{minipage}
\vspace{-3mm}
\end{figure*}

Algorithm~\ref{alg:maqr_main} is a specific implementation of a
more general model-agnostic algorithm, in which we directly estimate conditional
quantiles from the data with tools from conditional density estimation.
We note that using KDEs for conditional density estimation is a well studied problem with theoretical guarantees \citep{holmes2012fast, hyndman1996estimating, stute1986almost}.
In the case the distance in $\mathcal{X}$ is measured
using a uniform kernel with mild assumptions on the 
bandwidth, Algorithm~\ref{alg:maqr_main} falls under the guarantees stated by \citet{stute1986almost}.
\\

\noindent
{\bf Theorem 1}~\citep{stute1986almost}.
{\it Assume $\cy \subset \mathbb{R}$, $\mathcal{X} \subset \mathbb{R}^n$, $\text{dist}(x_i, x_j) := |{x_i - x_j}|_\infty$, and that $\hF_{\mathbf{E}|x}$ is constructed using the procedure given in line 5 of Algorithm~\ref{alg:maqr_main} (i.e. $x_i = x$). Further assume that, as $N \rightarrow \infty$, $d_N \rightarrow 0$ and that $\sum_{N\geq1}\exp(- \rho N d_{N}^{n}) < \infty$, $\forall \rho > 0$. Then, as $N \rightarrow \infty$, for almost all $x\in \mathcal{X}$, 
$\sup_{\eps} [\hF_{\mathbf{E}|x}(\eps) - \F_{\mathbf{E}|x}(\eps)] \rightarrow 0$ with probability 1.
}
\vspace{2mm}



This theorem states that in the limit of data,
for almost all $x \in \mathcal{X}$, 
the CDF estimate $\hF_{\mathbf{E}|x}$ will converge
uniformly to the true CDF $\F_{\mathbf{E}|x}$ with probability 1. 
The dataset, $D$, will therefore be populated
with 
good estimates of the
conditional quantile and quantile level pair for $x$.
In Appendix~\ref{app:maqr}, we state the general form of Algorithm~\ref{alg:maqr_main} and also demonstrate how the algorithm is model agnostic.
Through our experiments in Section~\ref{sec:experiments}, we will show empirically that utilizing these density estimates sidesteps the issues inherent to the pinball loss and produces much higher quality quantile predictions.

\vspace{-1mm}
\subsection{Explicitly balancing calibration and sharpness with the combined calibration loss} \label{sec:cali}
\vspace{-1mm}
While MAQR can produce strong results, its performance can suffer in high-dimensional settings, 
where nonparametric conditional density estimation methods falter.
Neural networks (NNs) have shown good performance in high
dimensional settings, given their high capacity to approximate
complex functions and recent advances in fast gradient-based optimization.
We therefore propose a loss-based approach to estimating conditional quantiles for NNs and other differentiable models. 

Drawing motivation from the \textit{arbitrary} balance between calibration and sharpness
that pinball loss \textit{implicitly} provides, 
we propose objectives separately for calibration and sharpness,
Then, we combine the two objectives into a single loss function that provides an 
explicit balance between calibration and sharpness that can be chosen by the end user.

We first consider calibration of a quantile prediction, $\hQ_p \in \mathcal{Y}$ 
for quantile level $p \in (0, 1)$.
Here, we omit conditioning on $x$ for clarity.
For this prediction to be average calibrated,
exactly a $p$ proportion of the true density 
should lie below $\hQ_p$, i.e. $\ppobs = P(Y \leq \hQ_p) = p$.
While calibration (e.g. $|\ppobs-p|$) is a non-differentiable objective,
by inducing a truncated distribution based on the current level of calibration, 
we can construct the following calibration objective, which is minimized
if and only if the prediction is average calibrated:
%
\begin{align}
    %
    \mathcal{C}(\hQ_p, p) 
    &= \mathbb{I}\{\hat{p}_p < p\} * \E[Y-\hQ_p | Y > \hQ_p] * P(Y > \hQ_p) \\ 
    & \hspace{5pt} + \mathbb{I}\{\hat{p}_p > p\} * \E[\hQ_p - Y | \hQ_p > Y] * P(\hQ_p > Y), 
    \text{where} \hspace{5pt} \hp_p = P(Y \leq \hQ_p). \nonumber
\end{align}
The empirical calibration objective, $\mathcal{C}(D, \hQ_p, p)$, is then defined as follows: 
\begin{align} \label{eq:emp_cal_obj}
    \begin{split}
        \mathcal{C}(D, \hQ, p) 
        & =\mathbb{I}\{\hp^{\text{obs}}_{avg} < p\} * \frac{1}{N}\sum_{i=1}^{N} \left[ (y_i - \hQ_p(x_i)) \mathbb{I}\{y_i > \hQ_p(x_i)\} \right] \\ 
        & \hspace{10pt} +\mathbb{I}\{\hp^{\text{obs}}_{avg} > p\} * \frac{1}{N}\sum_{i=1}^{N} \left[ (\hQ_p(x_i) - y_i) \mathbb{I}\{\hQ_p(x_i) > y_i\} \right].
    \end{split}
\end{align}


\textit{\textbf{Note 1:} Intuition of the calibration objective.}
For any given $p$, consider the case when the quantile estimate $\hQ_p$ is 
below the true $p\sth$ quantile $\mathbb{Q}_p$. 
Since $\hQ_p < \mathbb{Q}_p \Longrightarrow \hat{p}_p < p$, this implies that 
too much data density lies above $\hQ_p$.
In this case, $\mathcal{C}(\hQ_p, p)$ reduces to $\E[Y-\hQ_p | Y > \hQ_p] * P(Y > \hQ_p)$. $\hQ_p$ is pulled higher with the expectation of the truncated distribution that
places $\hQ_p$ at the lower bound of the support. 
In the opposite case, when $\hQ_p > \mathbb{Q}_p$, $\hQ_p$ is pulled lower 
by the same logic.

\textit{\textbf{Note 2:} Is the proposed calibration objective a proper scoring rule?}
Strictly speaking, the calibration objective is a non-decomposable function,
hence deviates from the standard convention of proper scoring
rules \cite{gneiting2007strictly}, which
can be ``decomposed'' into scores for individual examples $(x_i, y_i)$.
This simply arises from the fact that
measuring average calibration (i.e. $\pobs$) is non-decomposable.
Proper scoring rules are defined such that an optimum of the \textit{expected score} (or \textit{risk}, if we consider the score as a \textit{loss function}) occurs at the true distribution quantity.
While an example level \textit{loss} or \textit{score} does
not exist due to non-decomposability, 
we can still show the (expectation-level) score
(i.e. $\mathcal{C}(\hQ_p, p)$) is minimized by the true
distribution and hence enjoys the optimum property of proper scoring rules.

{\bf Proposition 2}. \label{prop:cal_obj}
{\it For any quantile level $p \in (0, 1)$, 
the true quantile function $\mathbb{Q}_p$ minimizes the calibration objective, 
$\mathcal{C}(\hat{\mathbb{Q}}_p, p)$.
Further, on a finite dataset $D$, the empirical calibration objective, $\mathcal{C}(D, \hat{\mathbb{Q}}_p, p)$,
is minimized by an average calibrated solution on $D$, i.e. when $\pobs(D, p) = p$.}


\textbf{Proof:} The proof is given in Appendix~\ref{app:prop_2}.

\textit{\textbf{Note 3:} Non-zero gradients for miscalibrated predictions $\hQ_p$.}
We can further show that for a miscalibrated quantile prediction, 
the gradients of $\mathcal{C}$ are always non-zero. 
When $\hp_p < p$,
$\partial \mathcal{C}(\hQ_p, p)/\partial \hQ_p$ $=$ $-P(Y > \hQ_p) < 0$.
Thus increasing $\hQ_p$
decreases the objective $\mathcal{C}$.
Similarly, when $\hp_p > p$,
$\partial \mathcal{C}(\hQ_p, p) / \partial \hQ_p$
$=$ $P(Y < \hQ_p) > 0$, and an analogous argument follows (proof in Appendix~\ref{app:note_3}).



As discussed in Section~\ref{sec:metrics}, average calibration by itself is not 
a sufficient condition for meaningful UQ, 
hence we also
desire \textit{sharp} quantile models, with more-concentrated (less dispersed) distributions.
We can induce this property in quantile predictions by 
predicting the $(1-p)$\textsuperscript{th} quantile $\hQ_{1-p}(x_i)$ alongside each
prediction $\hQ_p(x_i)$ and penalizing the width between the quantile predictions:
\begin{align}
    \mathcal{P}(\hQ_p, p) = \E \left[ \left| \hQ_p - \hQ_{1-p} \right| \right].
\end{align}
The empirical sharpness objective, $\mathcal{P}(D, \hQ_p, p)$, is then defined as follows: 
\begin{align} \label{eq:emp_sharp_obj}
    &\mathcal{P}(D, \hQ, p) = \frac{1}{N}\sum_{i=1}^{N} 
        \begin{cases} 
            \hQ_{1-p}(x_i) - \hQ_p(x_i) \hspace{5pt} (p \leq 0.5)\\
            \hQ_p(x_i) - \hQ_{1-p}(x_i) \hspace{5pt} (p > 0.5).
        \end{cases}
\end{align}
It is important to note that the true underlying distribution will not have $0$ sharpness if there is significant noise, and sharpness should be optimized subject to calibration.
Therefore, we should only penalize sharpness when the data suggests our quantiles are too dispersed, 
i.e. when $\left| p^{\text{obs}}_{avg}(p) - p^{\text{obs}}_{avg}(1-p) \right|$, the \textit{observed coverage} between the pair of quantiles $\hQ_p(x_i)$ and $\hQ_{1-p}(x_i)$,
is greater than $ \left| 2p-1 \right|$, the \textit{expected coverage}.

Combining the calibration and sharpness terms, we have the \textbf{combined calibration loss}
\begin{align}
    \label{eq:combined-calibration-loss}
    \mathcal{L}(D, \hQ_p, p) = (1-\lambda)\mathcal{C}(D, \hQ_p, p) + \lambda \mathcal{P}(D, \hQ_p, p).
\end{align} 
The hyperparameter $\lambda \in [0, 1]$ sets the explicit balance between calibration and sharpness. 
Note that setting $\lambda = 0$ may not always be desirable, since optimizing $\mathcal{C}(D, \hQ_p, p)$ alone may converge to quantiles of the marginal distribution, $\marginal$.
Further, in certain downstream applications that utilize UQ,
a sharper prediction, even at the cost of worse calibration, 
may result in higher utility, and $\lambda$ can be tuned according to the utility function of the application.
In our experiments, we tune $\lambda$ by cross-validating with adversarial group calibration as it is the strictest notion of 
calibration that can be estimated with a finite dataset.
Since we learn a quantile model that outputs the conditional quantile 
estimates for all probabilities, our training objective is
$\E_{p\sim \text{Unif}(0, 1)}\mathcal{L}(D, \hat{\Q}_p, p)$.






\vspace{-1mm}
\subsection{Encouraging calibration of centered intervals with the interval score}
\label{sec:int}
\vspace{-2mm}
The combined calibration loss (Eq.~\ref{eq:combined-calibration-loss}) optimizes
average calibration, which targets observed probabilities below a quantile.
In many applications, however, we may desire a centered prediction interval (PI)
which requires a pair of quantile predictions. A centered $95\%$ PI, for example,
is a pair of quantile predictions at quantile levels $0.025$ and $0.975$.
Hence, for the average calibration of the $p\sth$ \textit{centered interval}, we want
$\left[\pobs(0.5+\frac{p}{2}) - \pobs(0.5-\frac{p}{2})\right]$
(the PI's observed probability, a.k.a. prediction interval coverage probability (PICP)
\citep{tagasovska2019single, kabir2018neural, pearce2018high}) to be equal to the expected probability $p$.
While we can modify the objective in Eq.~\ref{eq:combined-calibration-loss} to adhere to this altered goal, 
here we propose simultaneously optimizing the \textbf{interval score} (or Winkler score) \citep{gneiting2007strictly, winkler1972decision} 
for all expected probabilities $p\in (0, 1)$, 
and bring to light a proper scoring rule that has largely been neglected for the purpose of \textit{learning quantiles}. 
While some previous works utilize the interval score to \textit{evaluate} interval predictions~\citep{bracher2021evaluating, bowman2020uncertainty, askanazi2018comparison, maciejowska2016probabilistic},
to the best of our knowledge, no previous work has focused on simultaneously optimizing it and
shown a thorough experimental evaluation as we provide in Section~\ref{sec:experiments}.

For a point $(x, y)$, if we denote a $(1-\alpha)$ centered PI as $\hat{l}, \hat{u}$,
i.e. $\hat{l} = \hQ_{\frac{\alpha}{2}}(x)$ and $\hat{u} = \hQ_{1-\frac{\alpha}{2}}(x)$,
the interval score is defined as 
$S_{\alpha}(\hat{l}, \hat{u}; y) = 
(\hat{u} - \hat{l}) + \frac{2}{\alpha}(\hat{l} - y)\mathbb{I}\{y < \hat{l}\} + \frac{2}{\alpha}(y - \hat{u})\mathbb{I}\{y > \hat{u}\}$.
We show in Appendix~\ref{app:int_score} that the minimum of the expectation of the interval score is attained at the
true conditional quantiles,
$\hat{l} = \Q_{\frac{\alpha}{2}}(\cdot)$,
$\hat{u} = \Q_{1 - \frac{\alpha}{2}}(\cdot)$.
We train our quantile model for all centered intervals (and hence all quantile levels)
simultaneously by setting our loss as
$\mathbb{E}_{\alpha \sim \text{Unif}(0,1)} S_{\alpha}$.

\vspace{-2mm}
\subsection{Inducing adversarial group calibration with group batching}
\label{sec:group_batch}
\vspace{-2mm}

The calibration loss (Section~\ref{sec:cali}) and the interval score (Section~\ref{sec:int}) optimize for the 
\textit{average} calibration of quantiles and centered intervals, respectively.
To get closer to individual calibration, 
one condition we can additionally require is \textit{adversarial group calibration}.
Since adversarial group calibration requires average calibration over any subset of non-zero measure over the domain, 
this is not fully observable with finite datasets $D$ for all subset sizes.
However, for any subset in $D$ with enough datapoints, we can still estimate average calibration over the subset.
Hence, we can apply our optimization objectives onto appropriately large subsets 
to induce adversarial group calibration.

In practice, this involves constructing subsets within the domain and 
taking gradient steps based on the loss over each subset.
In naive implementations of stochastic gradient descent,
a random batch is drawn \textit{uniformly} from the 
training dataset $D$, and a gradient step is taken according to the loss over
this batch. This is also the case in \textit{SQR} \citep{tagasovska2019single}.
The uniform draw of the batch will tend to preserve $\F_{\bX}$ (the marginal distribution of $\bX$),
hence optimizing average calibration over this batch will only induce average calibration of the model.

Instead, deliberately grouping the datapoints based on input features, and then batching
and taking gradient steps based on these batches, induces 
better adversarial group calibration.
We find in our experiments that adversarial group calibration improves significantly
with simple implementations of group batching,
and in Section~\ref{sec:ablation}, we show through an ablation study 
that group batching can improve average calibration and
adversarial group calibration of \textit{SQR} as well.

To summarize, the main idea we introduce here with group batching is that, 
only taking uniform batches from the training set 
(thus only drawing batches which preserve $\F_{\textbf{X}}$) 
can be detrimental when optimizing for calibration. 
Thus, additionally drawing batches based on deliberate groupings within the training set (thus, batches which do not preserve $\F_{\textbf{X}}$) can help to induce a stronger notion
of calibration (i.e. adversarial group calibration) in the model than average calibration.
This concept is quite general and allows for variations in implementations when 
constructing the groups. 
In Appendix~\ref{app:group_batching} we provide details on how we implemented group batching for our experiments and ablation study.




%% file: experiments.tex
\vspace{-1mm}
\section{Experiments}
\label{sec:experiments}
\vspace{-1mm}

\vspace{-2mm}

\begin{figure*}[t!]
    \centering
    
    \includegraphics[width=\textwidth]{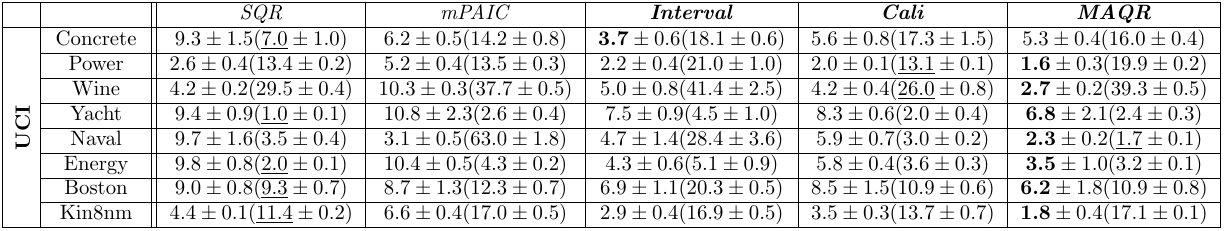}
    \includegraphics[width=\textwidth]{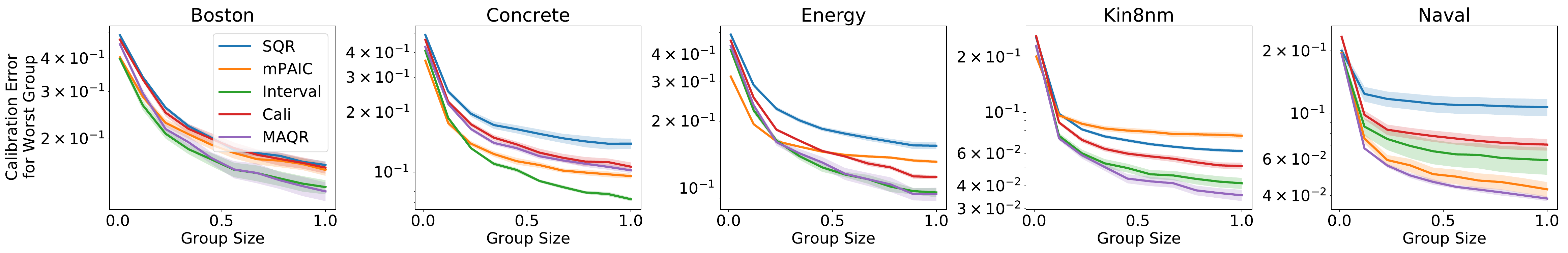}
    
    \caption{\textbf{UCI Experiments.}  
    \textbf{(Top Table):} Average calibration (measured by ECE) and sharpness in parentheses. The best mean ECE for each dataset has been bolded and the best mean sharpness has been underlined (all values multiplied by 100 for readability).
    \textbf{(Bottom Figure):} Adversarial group calibration for the first 5 UCI datasets (full set of results in Appendix~\ref{app:all_uci_results}). Group size refers to proportion of test dataset size. 
    }
    \label{fig:uci_table_figure}
    \vspace{-1mm}
\end{figure*}

\begin{figure*}[t!]
    \centering
    
    \includegraphics[width=0.95\textwidth]{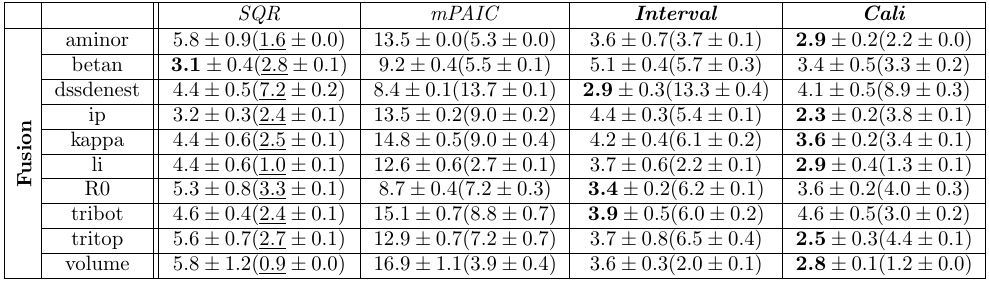}
    \includegraphics[width=\textwidth]{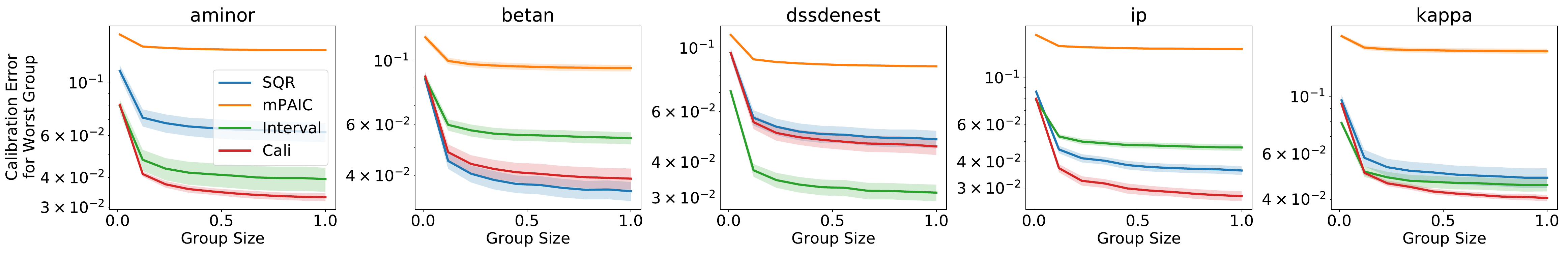}
    
    \caption{\textbf{Fusion Experiments.}  
    \textbf{(Top Table):} Average calibration (measured by ECE) and sharpness in parentheses. The best mean ECE for each dataset has been bolded and the best mean sharpness has been underlined (all values multiplied by 100 for readability).
    \textbf{(Bottom Figure):} Adversarial group calibration for the first 5 fusion datasets (full set of results in Appendix~\ref{app:all_fusion_results}). Group size refers to proportion of test dataset size. 
    }
    \label{fig:fusion_table_figure}
    \vspace{-1mm}
\end{figure*}

We demonstrate the performances of our proposed methods on the standard 8 UCI datasets \cite{asuncion2007uci}, 
and on a real-world problem in nuclear fusion. 
To assess the predictions, we use all metrics from Section~\ref{sec:metrics} that can be estimated with a finite test dataset: 1) average calibration vs sharpness, 2) adversarial group calibration, 
3) centered interval calibration, 4) check score, and 5) interval score.
The results for the first two of these metrics are displayed in the main body of the paper, and the results for the other three metrics are shown in Appendix~\ref{app:all_experiments} due to space restrictions.
We describe how each of these evaluation metrics are calculated in
Appendix~\ref{app:measuring_metrics}.

We provide comparisons against current state-of-the-art UQ methods 
for which computing the above metrics is tractable. 
\textit{SQR} \cite{tagasovska2019single} is an NN model that optimizes the pinball loss
for a batch of random quantile levels $p\sim \text{Unif}(0, 1)$.
\textit{mPAIC} \cite{zhao2020individual} is an extension of 
probabilistic neural networks \cite{lakshminarayanan2017simple, nix1994estimating}
that optimizes a combination of the standard Gaussian NLL and 
a loss that induces individual calibration.
While additional quantile and PI based UQ methods exist, 
they are mostly designed to output a single quantile level or interval coverage level,
which makes measuring calibration extremely expensive (training up to 100 models separately).
For these methods, we provide a comparison on a simplified task of predicting the $95\%$ PI in Appendix~\ref{app:coverage_95}.
Lastly,
we could also consider the recalibration algorithm of \citet{kuleshov2018accurate},
which is not a standalone UQ method, but a post-hoc refinement step that can be applied on top of other methods. 
We discuss recalibration results in Appendix~\ref{app:recal}.
%
All results report the mean and error across 5 trials. Error bars and shaded bands in plots indicate $\pm 1$ standard error.

\vspace{-3mm}
\subsection{UCI and Fusion Experiments} \label{sec:uci_fusion}
\vspace{-2mm}

\begin{figure*}[t] 
\begin{center} 
\includegraphics[width=0.98\textwidth]{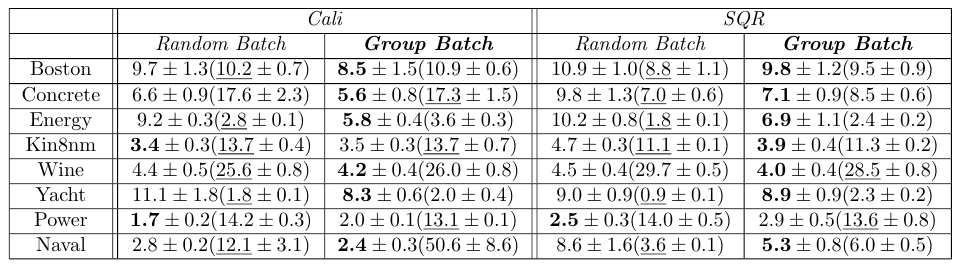}
\caption{\textbf{Group Batching Ablation: Average Calibration and Sharpness.}
    The table shows mean ECE and sharpness (in parentheses) and their standard error with and without group batching. The best mean ECE for each dataset has been bolded and the best mean sharpness has been underlined for \textit{Cali} and \textit{SQR} separately. All values have been multiplied by 100 for readability.} \label{fig:group_batch_ablation_main}
\end{center}
\vspace{-5mm}
\end{figure*}

\textbf{UCI datasets:}
We evaluate 5 methods on the 8 UCI benchmark datasets: 
three proposed algorithms---\textit{MAQR}, \textit{Cali} (combined calibration loss), and \textit{Interval} (interval score)---and 2 alternative algorithms---\textit{SQR} and \textit{mPAIC}.
Appendix~\ref{app:uci_exp_details} includes more details on the experiment setup and hyperparameters.

\textbf{Fusion datasets:}
We further evaluate the methods on a 
high-dimensional task from nuclear fusion: quantifying uncertainty in plasma dynamics. 
Recently, there has been increasing interest in applying machine learning
to prediction and control tasks in the area of nuclear
fusion \cite{chung2020offline, char2019offline, mehta2020neural}.
The plasma data in this paper was recorded from the DIII-D tokamak, a nuclear fusion device operated by General Atomics \cite{luxon2002design}. 
Plasma dynamics during fusion reactions are highly stochastic and running live fusion experiments is costly. 
Hence, practitioners use this dataset to learn a dynamics model of the system
for various purposes, such as controller learning to optimize reaction efficiency and stability \cite{fu2020machine, boyer2019feedback, boyer2019real}.
There are 10 scalar target signals to model in this dataset (full list in Appendix~\ref{app:fusion_exp_details}), 
each describing a particular aspect of the current state of plasma.
For each signal, the input features are 468 dimensional.
We do not apply \textit{MAQR} on this dataset because of the high computational costs and statistical challenge associated with nonparametric density estimation in high dimensions.
Appendix~\ref{app:fusion_exp_details} provides more details on the dataset, experiment setup, and hyperparameters.

\textbf{Analysis of Results:}
Figure~\ref{fig:uci_table_figure} displays average calibration-sharpness for all 8 UCI datasets and
adversarial group calibration for the first 5 UCI datasets (alphabetical order).
\textit{MAQR} produces the best average calibrated models on 7 of the 8 UCI datasets, 
and adversarial group calibration also indicates that \textit{MAQR} 
tends to achieve the lowest calibration error across 
\textit{any} random subgroup of \textit{any} size with more than one point (6 out of 8 datasets).
Excluding \textit{MAQR}, \textit{Interval} and \textit{Cali} achieve competitive average calibration and adversarial group calibration on 7 out of 8 UCI datasets.
Notably, \textit{SQR} tends to produce the sharpest predictions across all 
datasets (often at the cost of worse calibration), 
which is not surprising following the discussion in Section~\ref{sec:metrics}. 
Lastly, we also note that \textit{Interval} and \textit{Cali}  
tend to be
less brittle compared to \textit{SQR} and \textit{mPAIC}, 
which incur major failures in some cases (e.g. \textit{mPAIC} in Kin8nm, \textit{SQR} in Naval).

The fusion experiment results display a similar pattern (Figure~\ref{fig:fusion_table_figure}).
\textit{Interval} and \textit{Cali} achieve competitive average
calibration and adversarial group calibration on 9 out of 10 fusion datasets, 
and \textit{SQR} produces the sharpest UQ at the cost worse average and adversarial group calibration (except on betan).
We also observe that \textit{mPAIC} performs the poorest on the fusion datasets, with least calibrated and least sharp predictions.
It is generally known that plasma dynamics display complex stochasticity
\cite{fu2020machine, kowal2020kelvin}, and hence we suspect \textit{mPAIC}'s
performance degrades significantly because 
it assumes a Gaussian output and is trained according to the Gaussian likelihood.

Only a subset of the plots and metrics are shown in the main paper due to space restrictions. Readers are encouraged to see the full set of results in Appendix~\ref{app:all_experiments}, but we summarize the main findings here.
The check score, interval score, and centered interval calibration
in the tables in Appendix~\ref{app:all_uci_results} also rank 
\textit{MAQR}'s prediction as best on the UCI datasets.
This is surprising since \textit{SQR} and \textit{Interval} train NNs of the same capacity to \textit{explicitly minimize} the check and interval scores, respectively.
This indicates the distribution predicted by \textit{MAQR} is fundamentally different 
as it utilizes direct estimates of the conditional distribution,
while the other methods all optimize a specific loss function.
The centered interval calibration metric on the fusion datasets (Appendix~\ref{app:all_fusion_results}) 
indicates that \textit{Cali} and \textit{Interval} both
produce competitive centered PI's (9 out of 10 fusion datasets), 
suggesting that both methods have generally estimated the conditional quantiles better than the alternative algorithms.

\vspace{-2mm}
\subsection{Ablation Study: Effect of Group Batching} \label{sec:ablation}
\vspace{-2mm}
We provide an ablation study on the effect of group batching (experiment details in Appendix~\ref{app:ablation_study_exp_setting}).
Figure~\ref{fig:group_batch_ablation_main} displays how group batching 
affects average calibration-sharpness of
\textit{Cali} and \textit{SQR} on all 8 UCI datasets.
Average calibration improved on 6 out of 8 datasets for \textit{Cali},
and on 7 out of 8 datasets on \textit{SQR}.
While sharpness tends to worsen with group batching, 
it's important to note that less sharp (i.e. more dispersed) predictions 
\textit{are} desirable if there is high noise 
in the true data distribution.
In Appendix~\ref{app:full_ablation_results}, we show that group batching also improves adversarial group calibration, suggesting that the underlying data distribution truly does have high noise.
This study indicates that deliberately taking batches during training that do not follow $\F_{\mathbf{X}}$ can improve UQ performance.



%% file: conclusion.tex
\vspace{-1mm}
\section{Discussion}
\label{sec:discussion}
\vspace{-1mm}
In this paper, we have proposed four methods to improve quantile estimates 
for calibrated uncertainty quantification in regression.
We assert that the pinball loss may not be an adequate objective
to optimize in order to achieve calibration,
and that our proposed methods provide better means of learning 
calibrated conditional quantiles.
We have also extended the scope of regression models on which quantile-based UQ 
can be applied by developing a model-agnostic method.
This can be of practical interest 
to users that have specific training infrastructure or preexisting regression procedures, 
since these procedures can be leveraged to quantify uncertainty without much additional overhead.

By providing an extensive evaluation with a suite of metrics,
we also aim to show
that there is not one single metric that captures full information about 
the performance of a UQ procedure. Rather, a holistic review of multiple types of metrics
sheds light on different aspects of performance.
This point has motivated us to additionally develop and make publicly available 
\href{https://github.com/uncertainty-toolbox/uncertainty-toolbox}{\textit{Uncertainty Toolbox}} 
\cite{chung2021uncertainty}, a Python library for evaluation, visualization and
recalibration of predictive uncertainty, which includes the full suite of metrics
discussed in this work.

In certain applications, users may also be concerned with modeling epistemic uncertainty.
This is somewhat orthogonal to the goals of this paper;  nevertheless, we provide a
discussion on how bootstrapped ensembles of our methods can incorporate epistemic
uncertainty in Appendix~\ref{app:epistemic}.

%% file: acknowledgments.tex
\subsection*{Acknowledgments}

This work was funded in part by DOE grant number DE-SC0021414 and DE-AC02-76SF00515.

Willie Neiswanger was supported in part by NSF (\#1651565), ONR (N000141912145), AFOSR (FA95501910024), ARO (W911NF-21-1-0125), DOE (DE-AC02-76SF00515) and Sloan Fellowship. 

Ian Char was also supported by NSF grant DGE1745016. Any opinions, findings, and conclusions or recommendations expressed in this material are those of the author(s) and do not necessarily reflect the views of the National Science Foundation.


%% file: appendix.tex
\newpage
\appendix
\onecolumn

\toptitlebar
\begin{adjustwidth}{3mm}{3mm}
\vspace{-2mm}
{\LARGE\bf Appendix for ``Beyond Pinball Loss: Quantile Methods for Calibrated Uncertainty Quantification''}
\vspace{-1mm}
\end{adjustwidth}
\bottomtitlebar
\vspace{3mm}


\label{sec:appendix}

\input{app_theory}
\newpage
\input{app_maqr}

\newpage
\input{app_exp_details}

\newpage
\input{app_exp_results}
\clearpage
\newpage
\input{app_add_exp}
\clearpage
\newpage
\input{app_ablation}

\clearpage
\newpage
\input{app_epistemic}
\newpage
\input{app_add_discussion}

\clearpage

%% file: app_theory.tex
\section{Theoretical Results}
\subsection{Proof of Proposition 1} \label{app:prop_1}

{\bf Proposition 1}. \label{prop:pinball}
{\it Consider a finite dataset $D$, the pinball loss $\rho_\tau$ (Eq.~\ref{eq:pinball_loss}) and a quantile model
$f: \mathcal{X} \times (0, 1) \rightarrow \mathcal{Y}$ that is average calibrated on $D$, i.e. $\text{ECE}(D, f) = 0$.
Then there always exists another quantile model $g: \mathcal{X} \times (0, 1) \rightarrow \mathcal{Y}$, such that, for any quantile level $\tau \in (0, 1)$, 
$g$ has lower pinball loss than $f$ on $D$, i.e. 
$\sum_{i=1}^{N}\rho_\tau(y_i, g_\tau(x_i)) < \sum_{i=1}^{N}\rho_\tau(y_i, f_\tau(x_i))$,
but worse average calibration than $f$, i.e. $\text{ECE}(D, g) > \text{ECE}(D, f)$.} 

\textbf{Proof:}

Denote $\{p_j\}_{j=1}^{m}$ 
as the set of quantiles used to compute $ECE(D,f)$.
Since $f$ is average calibrated on $D$,
for any $\tau \in \{p_j\}_{j=1}^{m}$, $\pobs(D, \tau)$ of $f$ is equal to $\tau$, i.e. $\frac{1}{N}\sum_{i=1}^{N} \mathbb{I}\{y_i \leq f_\tau(x_i)\} = \tau$. 

Hence,
\begin{itemize}
    \item $|\{(x_i, y_i) \mid y_i \leq f_{\tau}(x_i)\}| = \tau*N$, 
    \item $|\{(x_i, y_i) \mid y_i > f_{\tau}(x_i)\}| = (1-\tau)*N$.
\end{itemize}

Denote $\{(x_i, y_i) \mid y_i \leq f_{\tau}(x_i)\}$ as $S^{\text{under}}_{f}$
and $\{(x_i, y_i) \mid y_i > f_{\tau}(x_i)\}$ as $S^{\text{over}}_{f}$.

Let $\rho_\tau(D, f_{\tau})$ be the pinball loss of $f_\tau$ on $D$, i.e. $\rho_{\tau}(D, f_{\tau}) = \sum_{i=1}^N \rho_\tau(y_i, f_\tau(x_i))$.

Then, 
\begin{align*}
\begin{aligned}
\rho_\tau(D, f_{\tau})
= \sum_{i=1}^{N}\rho_\tau(y_i, f_\tau(x_i))
&= \sum_{(x_i,y_i) \in S^{\text{under}}_{f}} (f_\tau(x_i) - y_i)(1-\tau) \\
&+ \sum_{(x_i,y_i) \in S^{\text{over}}_{f}} (f_\tau(x_i) - y_i)(-\tau).
\end{aligned}
\end{align*}

We can construct another quantile model $g$, 
s.t. its prediction for $\tau$, $g_\tau$, is as follows:
take any point $(x_k, y_k) \in S^{\text{over}}_{f}$
and set $g_\tau(x_k)$ 
$= y_k - \frac{\tau}{2(1-\tau)}(f_{\tau}(x_k) - y_k)$. 
For all other points, $(x_i, y_i)_{i\neq k}\in D$, set $g_{\tau}(x_i) = f_{\tau}(x_i)$.


Since $y_k - g_{\tau}(x_k) = \frac{\tau}{2(1-\tau)}(f_{\tau}(x_k) - y_k) < 0$,
we have that $y_k < g_{\tau}(x_k)$.

Therefore, 
$|\{(x_i, y_i) \mid y_i \leq g_{\tau}(x_i)\}| = (\tau*N) + 1$ and 
$|\{(x_i, y_i) \mid y_i > g_{\tau}(x_i)\}| = ((1-\tau)*N) - 1$.

Therefore, $|\pobs(D, \tau)$ of $g$ $-$ $\tau |$ $>$ $|\pobs(D, \tau)$ of $f$ $-$ $\tau |$, which implies that according to $\{p_j\}_{j=1}^{m}$, 
$ECE(D, g) > ECE(D, f)$.

We further consider the pinball loss of $g_{\tau}$ on $D$, $\rho_\tau(D, g_\tau)$.

$\rho_\tau(D, g_\tau) = \sum_{i=1}^{N}\rho_\tau(y_i, g_\tau(x_i)) =  \rho_\tau(y_k, g_\tau(x_k)) + \sum_{i\in[N], i\neq k}\rho_\tau(y_i, f_\tau(x_i))$.

Note that the term, $\rho_\tau(y_k, g_\tau(x_k))$ satisfies:
\begin{align*}
    \rho_\tau(y_k, g_\tau(x_k)) 
    &= (g_\tau(x_k) - y_k)(1-\tau) \\
    &= \frac{\tau}{2(1-\tau)}(y_k - f_{\tau}(x_k)) (1-\tau) \\
    &= \frac{\tau}{2}(y_k - f_{\tau}(x_k)) \\
    &< \tau(y_k - f_{\tau}(x_k)) \\
    &= (-\tau)(f_{\tau}(x_k) - y_k) \\
    &= \rho_\tau(y_k, f_\tau(x_k))   
\end{align*}

Therefore, $\rho_\tau(D, g_\tau) < \rho_\tau(D, f_\tau)$,
i.e. the pinball loss of $g_{\tau}$ on $D$ is less than the pinball loss of $f_{\tau}$ on $D$.

Note that for any $\tau \not\in \{p_j\}_{j=1}^{m}$, 
we can follow the same steps to construct $g_\tau$ 
s.t. $|\pobs(D, \tau)$ of $g$ $-$ $\tau |$ $>$ $|\pobs(D, \tau)$ of $f$ $-$ $\tau |$, i.e. $g_\tau$ is more miscalibrated than $f_\tau$.
Therefore, for any quantile level $\tau \in (0, 1)$, $\sum_{i=1}^{N}\rho_\tau(y_i, g_\tau(x_i)) < \sum_{i=1}^{N}\rho_\tau(y_i, f_\tau(x_i))$, but
$\text{ECE}(D, g) > \text{ECE}(D, f)$. 

\hfill $\qedsymbol$
\vspace{5mm}

\subsection{Proof of Proposition 2} \label{app:prop_2}

{\bf Proposition 2}. \label{prop:cal_obj}
{\it For any quantile level $p \in (0, 1)$, 
the true quantile function $\mathbb{Q}_p$ minimizes the calibration objective, 
$\mathcal{C}(\hat{\mathbb{Q}}_p, p)$.
Further, on a finite dataset $D$, the empirical calibration objective, $\mathcal{C}(D, \hat{\mathbb{Q}}_p, p)$,
is minimized by an average calibrated solution on $D$, i.e. when $\pobs(D, p) = p$.}

\textbf{Proof:}\\
Recall the calibration objective for a quantile level $p\in(0,1)$,
\begin{align*}
    \mathcal{C}(\hQ_p, p) 
    &= \mathbb{I}\{\hat{p}_p < p\} * \E[Y-\hQ_p | Y > \hQ_p] * P(Y > \hQ_p) \\ 
    & \hspace{5pt} + \mathbb{I}\{\hat{p}_p > p\} * \E[\hQ_p - Y | \hQ_p > Y] * P(\hQ_p > Y), 
    \text{where} \hspace{5pt} \hp_p = P(Y \leq \hQ_p). \nonumber
\end{align*}

For the true quantile function $\Q_p$, $P(Y\leq \Q_p) = p$, thus achieves the minimum value of $0$ for $\mathcal{C}(\hQ_p, p)$, as
the two non-negative terms of $\mathcal{C}(\hQ_p, p) $ are $0$.

Further, recall the empirical calibration objective,
\begin{align*}
    \begin{split}
        \mathcal{C}(D, \hQ, p) 
        & =\mathbb{I}\{\hp^{\text{obs}}_{avg} < p\} * \frac{1}{N}\sum_{i=1}^{N} \left[ (y_i - \hQ_p(x_i)) \mathbb{I}\{y_i > \hQ_p(x_i)\} \right] \\ 
        & \hspace{10pt} +\mathbb{I}\{\hp^{\text{obs}}_{avg} > p\} * \frac{1}{N}\sum_{i=1}^{N} \left[ (\hQ_p(x_i) - y_i) \mathbb{I}\{\hQ_p(x_i) > y_i\} \right] \\
    \end{split}
\end{align*}

An average calibrated solution on the dataset $D$ satisfies $\hp^{\text{obs}}_{avg} = p$, thus achieves the minimum value of $0$ for $\mathcal{C}(D, \hQ, p)$, as
the two non-negative terms of $\mathcal{C}(D, \hQ, p) $ are $0$. 

\hfill $\qedsymbol$
\vspace{5mm}

\subsection{Derivation of Gradients of Calibration Objective} \label{app:note_3}
Denote the CDF and PDF of the random variable $Y$ as $\mathbb{F}_{Y}$ and
$f_{Y}$.
\begin{itemize}
    \item When $\hp_p = P(Y \leq \hQ_p) < p$:\\
    \begin{align*}
        \mathcal{C}(\hQ_p, p) 
        &= \E[Y-\hQ_p | Y > \hQ_p] * P(Y > \hQ_p)\\
        &= \left(\E[Y | Y > \hQ_p]-\hQ_p\right) * P(Y > \hQ_p)\\
        &= \left(\frac{\int_{\hQ_p}^{\infty} y f_Y(y) dy}{P(Y > \hQ_p)} -\hQ_p\right) * P(Y > \hQ_p)\\
        &= \left(\int_{\hQ_p}^{\infty} y f_Y(y) dy \right)
        - \left(\hQ_p * P(Y > \hQ_p)\right) \\
        &= \left(\int_{\hQ_p}^{\infty} y f_Y(y) dy \right)
        - \left(\hQ_p * (1 - F_Y(\hQ_p))\right) \\
    \end{align*}
    
    Note that,
    \begin{align*}
        \frac{\partial \left(\int_{\hQ_p}^{\infty} y f_Y(y) dy \right)}{\partial \hQ_p} 
        &= -\hQ_p * f_Y(\hQ_p) \\ 
        \frac{\partial  \left(\hQ_p * (1 - F_Y(\hQ_p)\right) }{\partial \hQ_p}
        &= (1 - F_Y(\hQ_p) + \hQ_P * (-f_y(\hQ_p))
    \end{align*}
    
    Therefore,
    \begin{align*}
        \frac{\partial \mathcal{C}(\hQ_p, p)}{\partial \hQ_p} 
        &= \frac{\partial \left(\int_{\hQ_p}^{\infty} y f_Y(y) dy \right)}{\partial \hQ_p} - \frac{\partial  \left(\hQ_p * (1 - F_Y(\hQ_p)\right) }{\partial \hQ_p} \\ 
        &= -\hQ_p * f_Y(\hQ_p) - (1 - F_Y(\hQ_p) + \hQ_P * f_y(\hQ_p) \\ 
        &= - (1 - F_Y(\hQ_p)\\
        &= -P(Y>\hQ_p)
    \end{align*}

    \item When $\hp_p = P(Y \leq \hQ_p) > p$:\\
    \begin{align*}
        \mathcal{C}(\hQ_p, p) 
        &= \E[\hQ_p - Y | \hQ_p > Y] * P(\hQ_p > Y)\\
        &= \left(\hQ_p - \E[Y | \hQ_p > Y]\right) * P(\hQ_p > Y)\\
        &= \left(\hQ_p - \frac{\int^{\hQ_p}_{-\infty} y f_Y(y) dy}{P( \hQ_p > Y)}\right) * P(\hQ_p > Y)\\
        &= \left(\hQ_p *P(\hQ_p > Y)\right)  - \left(\int^{\hQ_p}_{-\infty} y f_Y(y) dy \right)\\
        &= \left(\hQ_p * F_Y(\hQ_p)\right) - \left(\int^{\hQ_p}_{-\infty} y f_Y(y) dy \right)\\
    \end{align*}
    Note that,
    \begin{align*}
        \frac{\partial \left(\hQ_p * F_Y(\hQ_p)\right)}{\partial \hQ_p}
        &= F_Y(\hQ_p) + \hQ_P * f_Y(\hQ_p)\\
        \frac{\partial \left(\int^{\hQ_p}_{-\infty} y f_Y(y) dy \right)}{\partial \hQ_p} 
        &= \hQ_p * f_Y(\hQ_p)
    \end{align*}
    
    Therefore,
    \begin{align*}
        \frac{\partial \mathcal{C}(\hQ_p, p)}{\partial \hQ_p} 
        &= \frac{\partial  \left(\hQ_p * F_Y(\hQ_p\right) }{\partial \hQ_p} - 
        \frac{\partial \left(\int^{\hQ_p}_{-\infty} y f_Y(y) dy \right)}{\partial \hQ_p}\\ 
        &= F_Y(\hQ_p) + \hQ_P * f_Y(\hQ_p) - \hQ_p * f_Y(\hQ_p)\\
        &= F_Y(\hQ_p)\\
        &= P(Y<\hQ_p).
    \end{align*}
\end{itemize}
\hfill $\square$

\subsection{Optimum of Interval Score} \label{app:int_score}
Following notation from Section~\ref{sec:int}, we denote $\hat{l} = \hQ(x, \frac{\alpha}{2})$ and $\hat{u} = \hQ(x, 1-\frac{\alpha}{2})$, 
and we omit conditioning on $x$ for clarity.

Assume $\hat{l} \leq \hat{u}$. Then,
\begin{align*}
    &\mathbb{E}\left[ S_{\alpha}(\hat{l}, \hat{u}; y) \right]\\
    &= \int_{-\infty}^{\hat{l}} S_{\alpha}(\hat{l}, \hat{u}; y) d\F(y) 
    + \int_{\hat{l}}^{\hat{u}} S_{\alpha}(\hat{l}, \hat{u}; y) d\F(y) 
    + \int_{\hat{u}}^{\infty} S_{\alpha}(\hat{l}, \hat{u}; y) d\F(y)\\
    &= (\hat{u} - \hat{l})
    + \frac{2}{\alpha} \int_{-\infty}^{\hat{l}} (\hat{l} - y) d\F(y) 
    + \frac{2}{\alpha} \int_{\hat{u}}^{\infty} (y - \hat{u})  d\F(y)\\
    &\frac{\partial \mathbb{E}\left[ S_{\alpha}(\hat{l}, \hat{u}; y) \right]}{\partial \hat{l}} 
    = -1 + \frac{2}{\alpha}\int_{-\infty}^{\hat{l}} d\F(y) = -1 + \frac{2}{\alpha} \F(\hat{l})\\
    &\frac{\partial \mathbb{E}\left[ S_{\alpha}(\hat{l}, \hat{u}; y) \right]}{\partial \hat{u}} 
    = 1 - \frac{2}{\alpha}\int_{\hat{u}}^{\infty} d\F(y) = 1 - \frac{2}{\alpha}(1- \F(\hat{u})).
\end{align*}
Setting $\frac{\partial \mathbb{E}\left[ S_{\alpha}(\hat{l}, \hat{u}; y) \right]}{\partial \hat{l}}$ and $\frac{\partial \mathbb{E}\left[ S_{\alpha}(\hat{l}, \hat{u}; y) \right]}{\partial \hat{u}}$ to zero reveals the interval loss minima at the respective true quantiles,
\begin{align*}
    \F(\hat{l}) &= \frac{\alpha}{2} \text{  and  } \F(\hat{u}) = 1 - \frac{\alpha}{2}\\
    \text{i.e.  } \hat{l} &= \Q(\cdot, \frac{\alpha}{2}) \text{  and  } \hat{u} = \Q(\cdot, 1 - \frac{\alpha}{2})
\end{align*}
\hfill $\square$

%% file: app_maqr.tex
\section{Model Agnostic Quantile Regression} \label{app:maqr}
\subsection{General Algorithm for Model Agnostic Quantile Regression} 
As stated in Section~\ref{sec:maqr}, Algorithm~\ref{alg:maqr_main} is one implementation of a general 
model-agnostic quantile regression procedure, in which we take direct estimates of the target density
and regress onto these estimates. 
This general framework is stated in Algorithm~\ref{alg:general_mauq}

\begin{algorithm}[h ]
  \caption{General Algorithm for Model Agnostic Quantile Regression} 
\begin{algorithmic}[1]\label{alg:general_mauq}
    \STATE {\bfseries Input:} Train data $\{x_i, y_i\}_{i=1}^{N}$
    \STATE Initialize $D \leftarrow \varnothing$
    \FOR{$i=1$ {\bfseries to} $N$}
    \STATE Select a set of quantile levels $\{p_k\}_{k=1}^{m}$, \hspace{1mm}$p_k \in [0,1]$
    \STATE $\hat{q}_{i, p_k} \leftarrow$ KDE estimate of $\Q(x_i, p_k)$, $k=1,\ldots,m$
    \STATE $D \leftarrow D \cup \{x_i, p_k, \hat{q}_{i, p_k}\}_{k=1}^{m}$
    \ENDFOR 
    \STATE Use $D$ to fit a regression model $\hQ$ \\
    \hspace{10pt} $\hQ: (x_i, p_k) \mapsto \hat{q}_{i, p_k}$, \hspace{2mm}$k=1,\ldots,m$ 
  \STATE {\bfseries Output:} $\hat{g}$, \hspace{2mm}$k=1,\ldots,m$
\end{algorithmic}
\end{algorithm}

Algorithm~\ref{alg:maqr_main} implements the KDE step of Algorithm~\ref{alg:general_mauq} (Line 5 of Algorithm~\ref{alg:general_mauq})
by using a uniform kernel over $\cx$ (Line 2 of Algorithm~\ref{alg:CondQuantEst}) and $\cy$ (Lines 3,5,6 of Algorithm~\ref{alg:CondQuantEst}).

It should also be noted that many other conditional KDE methods can be used to construct the dataset $D$.
We refer the reader to \citet{holmes2012fast, hyndman1996estimating} for a more thorough treatment of methods in conditional KDE.

Lastly, this algorithm is model agnostic because \textit{any} regression model
can be used for $\hat{g}$ in Algorithm~\ref{alg:general_mauq}, and for $\hat{f}$ and $\hat{g}$ in Algorithm~\ref{alg:maqr_main}.
In our specific implementation of Algorithm~\ref{alg:maqr_main}, we used a neural network 
for $\hat{g}$ to fit the quantile dataset $D$, but we can also use other models, 
such as a random forest or gradient-boosted trees. 
In particular, we have replaced $\hat{g}$ in Algorithm~\ref{alg:maqr_main} with
a gradient-boosted tree model and observed very similar UQ performance on the UCI datasets
as reported in Section~\ref{sec:uci_fusion} (we omit numerical values because they are very similar 
and otherwise uninformative).

\subsection{Algorithm Complexity} \label{app:alg_complexity_analysis}

Lines 2 and 3 of Algorithm~\ref{alg:CondQuantEst} requires calculating the distance between $x_k$ and all other $x_i, 1\leq i \leq N$, 
and ordering these distances to construct the empirical CDF. 
Let the distance calculation between a pair of points take constant $C$ time. 
Ordering the distance requires sorting the $N$ distances. 
Hence, Lines 5 and 6 takes $\mathcal{O}(N\log N)$ time.

If $K$ points are in the set $E_{k,d_{N}}$, 
Lines 5, 6, 7 of Algorithm~\ref{alg:CondQuantEst} are done for each of the $K$ points.
We consider the worst case when each set $E_{k,d_{N}}$ contains all $N$ points, 
which costs $\mathcal{O}(N)$.

The above two procedures are done for all $N$ points (Line 4 of Algorithm~\ref{alg:maqr_main}), 
therefore the for loop from Lines 4 to 7 in Algorithm~\ref{alg:maqr_main} requires $\mathcal{O}(N^2 \log N)$ time. 
This loop takes into account creating the dataset $D$ for the quantile model. 
The rest of the algorithm constitutes fitting a regression model with this dataset $D$, which we do not analyze here.

We now discuss the space complexity. 
In Lines 5, 6, 7 of Algorithm~\ref{alg:CondQuantEst}, we only draw the quantiles at which a discontinuous step occurs in the constructed empirical CDF. 
For example, if we constructed an empirical CDF with three equally weighted points, we will only draw the quantiles $[1/3, 2/3, 1]$.
Following this procedure, in the worst case, for each of the $N$ points, we will construct a CDF with all $N$ points, and hence draw $N$ quantiles to append to the quantile model dataset $D$. 
If we consider the space complexity of the mean function and the quantile model to be constant, 
the algorithm requires $O(N^2)$ space in total.

%% file: app_exp_details.tex
\section{Details on Datasets and Setup of Main Experiments} \label{app:data_exp_details}

\subsection{UCI Experiment Details} \label{app:uci_exp_details}
Here, we provide details on the setup of the UCI experiments presented in Section~\ref{sec:uci_fusion}.

For each of the 8 UCI datasets, we split $10\%$ of the data into the test set, and we further split $20\%$ of the remaining $90\%$ of the data for the validation, resulting in a train/validation/test split of proportions $72\%, 18\%, 10\%$.
For all tasks and all 8 datasets, the data was preprocessed by centering to zero mean and scaling to unit variance. 

We used the same NN architecture across all methods: 2 layers of 64 hidden units with ReLU non-linearities.
We used the same learning rate, $1e^{-3}$, and the same batch size, 64, for all methods. 
For all methods, training was stopped early if the validation loss did not decrease for more than $200$ epochs, until a maximum of $10000$ epochs.
If training was stopped early, the final model was backtracked to the model with lowest validation loss.

\textit{IndvCal} (individual calibration) has one hyperparameter, $\alpha$, which balances
the NLL loss and the individual calibration loss in the loss function.
We 5-fold cross-validated $\alpha$ in [0.0, 1.0] in 20 equi-spaced intervals based on  
Pareto optimality in test set NLL and adversarial group calibration.
If there were multiple $\alpha$ values that were Pareto optimal, we chose the value
that had the best test set adversarial group calibration.

\textit{Cali} (penalized calibration loss) has one hyperparameter, $\lambda$, which balances the calibration loss and sharpness penalty in the loss function.
We tuned $\lambda$ according to the same grid as above for $\alpha$, based on the criterion of adversarial group calibration.
Note that adversarial group calibration will not always favor lower values of $\lambda$
as $\lambda=0$ will only target average calibration, which, in the degenerate case, may converge
to the marginal distribution of $\F_{\bold{Y}}$. This state will achieve very poor adversarial group calibration. 

Group batching was applied to \textit{Interval} (interval score) and \textit{Cali} according to the implementation detailed in Appendix~\ref{app:group_batching}.
During training, we alternated between ``group batching epochs'' and ``regular batching epochs'' (where batches are drawn uniformly from the training set), 
and the frequency of group batching epochs was a hyperparameter we tuned with cross-validation in [1, 2, 3, 5, 10, 30, 100], based on the criterion of 
adversarial group calibration.

\textit{MAQR} has a two-step training process: we first learn a mean model, then construct a quantile dataset $D$, then regress onto this dataset with the quantile model. 
Both the mean model and the quantile model had the same NN architecture as mentioned above.
The mean model was trained with the MSE loss according to the same, aforementioned training procedure. 
For each UCI dataset, we learned one mean model from the first seed, and re-used this mean model for all other seeds.
Using this mean model, we then populated the quantile dataset according to the method outlined in Algorithm~\ref{alg:maqr_main}.
Algorithm~\ref{alg:maqr_main} requires one hyperparameter: the distance threshold in $\mathcal{X}$ space ($d_N$ in Line 2 of Algorithm~\ref{alg:CondQuantEst}). 
We tuned this hyperparameter by setting the minimum distance required to include $k$ number of points, on average, in constructing an empirical CDF at each training point. 
We tuned this parameter with cross-validation using the grid $k \in [10, 20, 30, 40, 50]$.
The quantile model was trained according to the same training procedure but with one difference: the batch size was set to 1024 because the quantile dataset $D$ could become very large due to many conditional quantile estimates at each training point.

All methods, for all datasets were repeated with 5 seeds: [0, 1, 2, 3, 4].

\subsection{Fusion Experiment Details} \label{app:fusion_exp_details}
We first describe the fusion dataset from Section~\ref{sec:uci_fusion}, then
the details of the fusion experiment set up.

The fusion dataset was recorded from the DIII-D tokamak in San Diego, CA, USA, and describes the dynamics of plasma during a nuclear fusion reaction within the tokamak.
Consent and access to use this dataset was obtained via
collaborations with the Princeton Plasma Physics Lab.

While the dataset in its raw format is a time-series of the state variables and action variables, for the purposes of a supervised learning problem to learn the dynamics of plasma, it has been re-structrued into a \textit{(state, action, next state)} format. 
Therefore, the modeling task at hand is to learn the mapping \textit{(state, action)} to \textit{(next state)}, and the UQ task is then to learn the distribution 
over the next state given the current state and action.

There are 10 plasma state variables that we use both as the input state variables and the target variables. For the action variables, we use the power level of 8 neutral beams, which are a primary means of controlling plasma in a tokamak. These variables are described in Figure~\ref{tab:fusion_variables}.

\begin{figure*} 
\begin{center}
\begin{tabular}{ |c|c| }
 \hline
            & \textbf{State Variables} \\\hline
 aminor     &  Minor Radius   \\ \hline
 dssdenest  &  Line Averaged Electron Density   \\ \hline
 efsbetan   &  Normalized Beta   \\ \hline
 efsli      &  Internal Inductance   \\ \hline
 efsvolume  &  Plasma Volume   \\ \hline
 ip         &  Plasma Curent   \\ \hline
 kappa      &  Elongation   \\ \hline
 R0         &  Major Radius   \\ \hline
 tribot     &  Bottom Triangularity   \\ \hline
 tritop     &  Top Triangularity   \\ \hline
\end{tabular}
\vspace{2mm}

\begin{tabular}{ |c|c| } 
\hline
            & \textbf{Action Variables} \\\hline
 pinj\_15l   &  Co-current Beam 1 Power   \\ \hline
 pinj\_15r   &  Co-current Beam 2 Power   \\ \hline
 pinj\_21l   &  Counter-current Beam 1 Power   \\ \hline
 pinj\_21r   &  Counter-current Beam 2 Power   \\ \hline
 pinj\_30l   &  Co-current Beam 3 Power   \\ \hline
 pinj\_30r   &  Co-current Beam 4 Power   \\ \hline
 pinj\_33l   &  Co-current Beam 5 Power   \\ \hline
 pinj\_33r   &  Co-current Beam 6 Power   \\ \hline
\end{tabular}  
\caption{\textbf{State and Action Variables for Fusion Dataset.} 10 variables describe the current state of plasma, and the action space is 8 dimensional.}
 \label{tab:fusion_variables}
\end{center}
\end{figure*}


As input to the dynamics model, we model the current state as a 200 millisecond (ms) history window of the 10 state variables and 8 action variables, 
and we model the current action as a 200ms window into the future of the 8 action variables. 
The target variables are modeled as the change (or delta) of the state variables 200ms in the future. 
The state and action features are engineered according to the method used by~\citet{fu2020machine}. 
Each 200ms window is taken as one ``frame'', the 200ms window is further divided into 2 ``frames'' of equal length (100ms each), as well as thirds. 
Then we calculate the mean, variance, and slope of each state and action variable for each of these frames, and collect these statistics as the features. 
Hence, each 200ms window for 1 variable is summarized into 18 features (3 statistics per frame, and 6 frames per 200ms window).
Since there are 10 state variables and 8 action variables, and since the state window is 200ms long and the action window is 200ms long, the input is a 468 dimensional array (10 state variables + 8 previous action variables + 8 current action variables for a total of 26 input variables, and 18 features per variable). 
Once these features are created, we centered and scaled the inputs and targets to zero mean and unit variance. 

There were a total of $100$K training data points, and $10$K validation and $10$K test data points.

The training and hyperparameter tuning procedures were exactly the same as 
the UCI experiments (detailed above in Appendix~\ref{app:uci_exp_details}), except for two differences: 1) we have increased the NN capacity to 3 hidden layers of 100 hidden units and 2) the batch size was set to 500.

The fusion experiment was likewise repeated 5 times with the seeds [0, 1, 2, 3, 4].

\subsection{Group Batching Implementation Details} \label{app:group_batching}
Group batching, as introduced in Section~\ref{sec:group_batch}, 
is a general procedure in which deliberate subsets of the training data
are constructed and batched from during train time. 
There is no ``correct'' method to form these subsets, because the main
point is to simply \textit{avoid drawing batches only from $\F_{\bX}$}.
One could consider thresholding the values of each dimension of the domain
and discretizing the subsets according to the thresholds, 
and also taking unions of these discretizations to form new subsets.

This implementation can be computationally demanding, 
because for each threshold setting, one has to make sure
to choose a subset with sufficiently many points, and iteratively
increase or decrease the dimension thresholds if no subset with
sufficiently many points can be found.
This computational cost increases significantly with dimension.

Our implementation of group batching for all our experiments were as follows: we sort the datapoints according to a single dimension, 
then take consecutive sets of size equal to the batch size, and use these 
sets as the batches to take gradient steps over during an epoch.
We repeat the above process by cycling through each dimension for sorting.
While this process is very simple and inexpensive and only considers 
a single dimension in constructing the subsets, 
this group batching scheme has shown to be very effective in our 
experiments.

\subsection{Calculation of Evaluation Metrics} \label{app:measuring_metrics}

To measure the calibration metrics (average, adversarial group, centered interval),
we discretized the expected probabilities from $0.01$ to $0.99$ in $0.01$ increments
(i.e. $0.01$, $0.02$, $\dots$, $0.97$, $0.98$, $0.99$)
and calculated ECE according to this finite discretization.

To measure centered interval calibration, for each expected probability $p$, 
we predict centered $100 \times p \%$ PIs, and calculate $\hat{p}^{\text{obs}}_{avg}$ 
as the proportion of test points falling within the PI, 
i.e. $\hat{p}^{\text{obs}}_{avg}(p)$ here would be calculated as 
\begin{align*}
    \hat{p}^{\text{obs}}_{avg}(p) \text{   for centered intervals   } = \frac{1}{N}\sum_{i=1}^{N} \mathbb{I}\{ \hQ_{(0.5-p/2)}(x_i) \leq y_i \leq \hQ_{(0.5 + p/2)}(x_i)\}.
\end{align*}

The procedure in which we measure adversarial group calibration is the following.
For a given test set, we scale group size between $1\%$ and $100\%$ of the full test set size, in 10 equi-spaced intervals, and for each group size, we draw 20 random groups from the test set and record the worst calibration incurred across these 20 random groups. This is also the method used by \citet{zhao2020individual} to measure adversarial group calibration.

Sharpness was measured as the mean width of the $95\%$ centered PI (i.e. between  $p=0.025$ and $0.975$). 

The proper scoring rules (check score, interval score) were measured as the average of the score on the test set.

\subsection{Discussion of Compute and Resources}
\label{app:compute}
All experiments were run on a single NVIDIA GeForce GTX 1080Ti GPU, with a Intel(R) Xeon(R) Silver 4110 CPU.
The fusion datasets were the largest (100K training) and highest dimensional (468 dimensional) and took the longest to run: training one model to convergence took roughly $\sim1$ hour.

%% file: app_exp_results.tex
\section{Full Experimental Results}
\label{app:all_experiments}

In this section, we provide all of the experimental results from 
Section~\ref{sec:experiments}, for all metrics:
1) average calibration - sharpness, 2) adversarial group calibration, 3) check score, 4) interval score, and 5) centered interval calibration

For average calibration and sharpness, we present the numeric tables
again, along with a visualization which plots calibration and sharpness
along the $x$ and $y$ axes.

\subsection{Full UCI Experiment Results}
\label{app:all_uci_results}

\begin{figure*}[ht!]
    \centering
    
    \includegraphics[width=\textwidth]{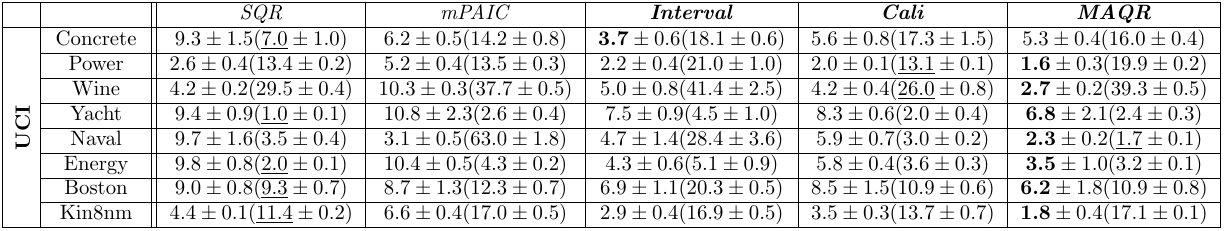}
    
    \caption{\textbf{UCI Average Calibration-Sharpness Table.}
    The table shows mean average calibration (measured by ECE) and sharpness in parentheses, along with $\pm 1$ standard error. 
    The best mean ECE for each dataset has been bolded and the best mean sharpness has been underlined. 
    All values have been multiplied by 100 for readability (same table as Figure~\ref{fig:uci_table_figure} (Top), repeated here for completeness).
    }
    \label{fig:uci_ece_sharp_table_full_app}
    \vspace{-1mm}
\end{figure*}

\begin{figure}[H]
    \centering
    \includegraphics[width=\textwidth]{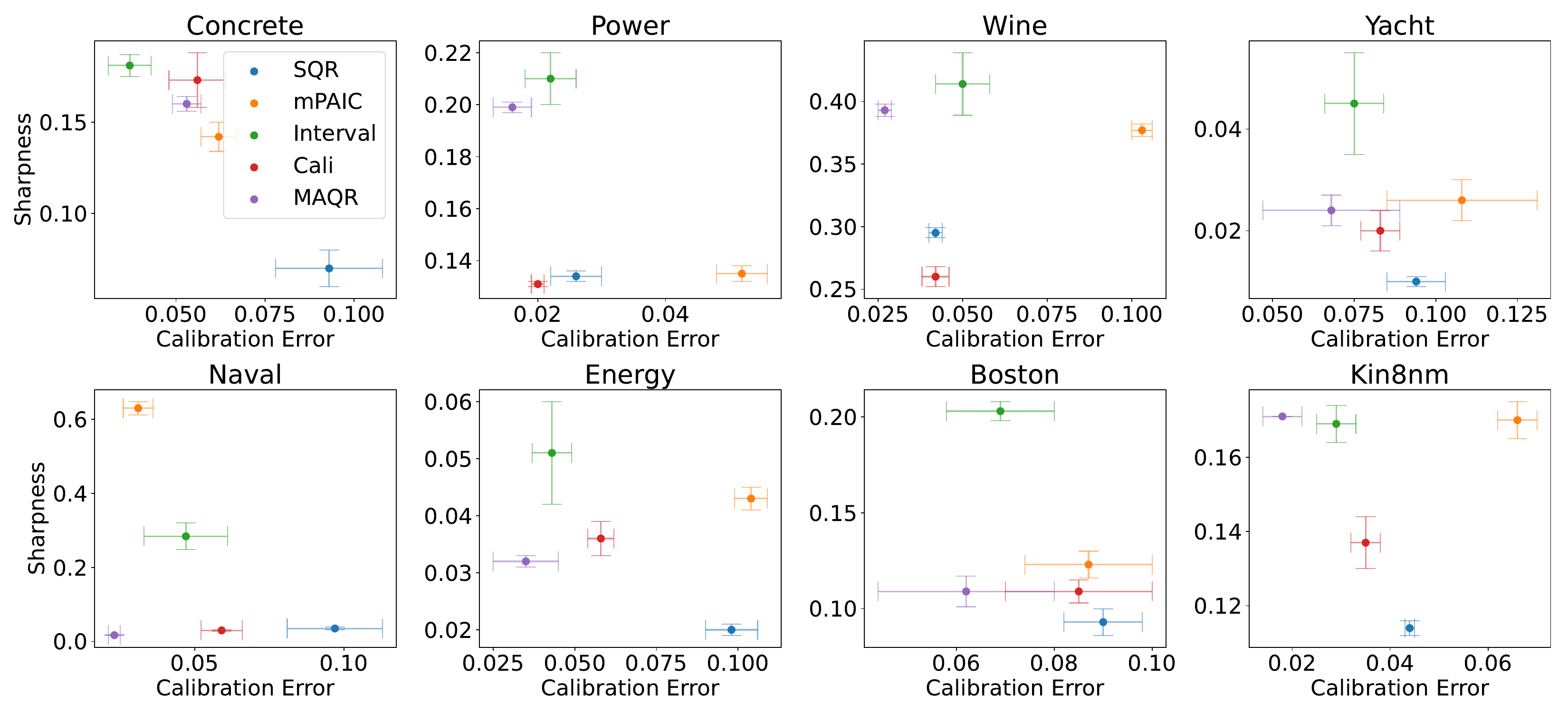}
    \caption{\textbf{UCI Average Calibration-Sharpness Plot.} Visualization of average calibration-sharpness from UCI experiments in Section~\ref{sec:uci_fusion}}
    \label{fig:full_uci_cal_sharp}
\end{figure}

\begin{figure}[H]
    \centering
    \includegraphics[width=\textwidth]{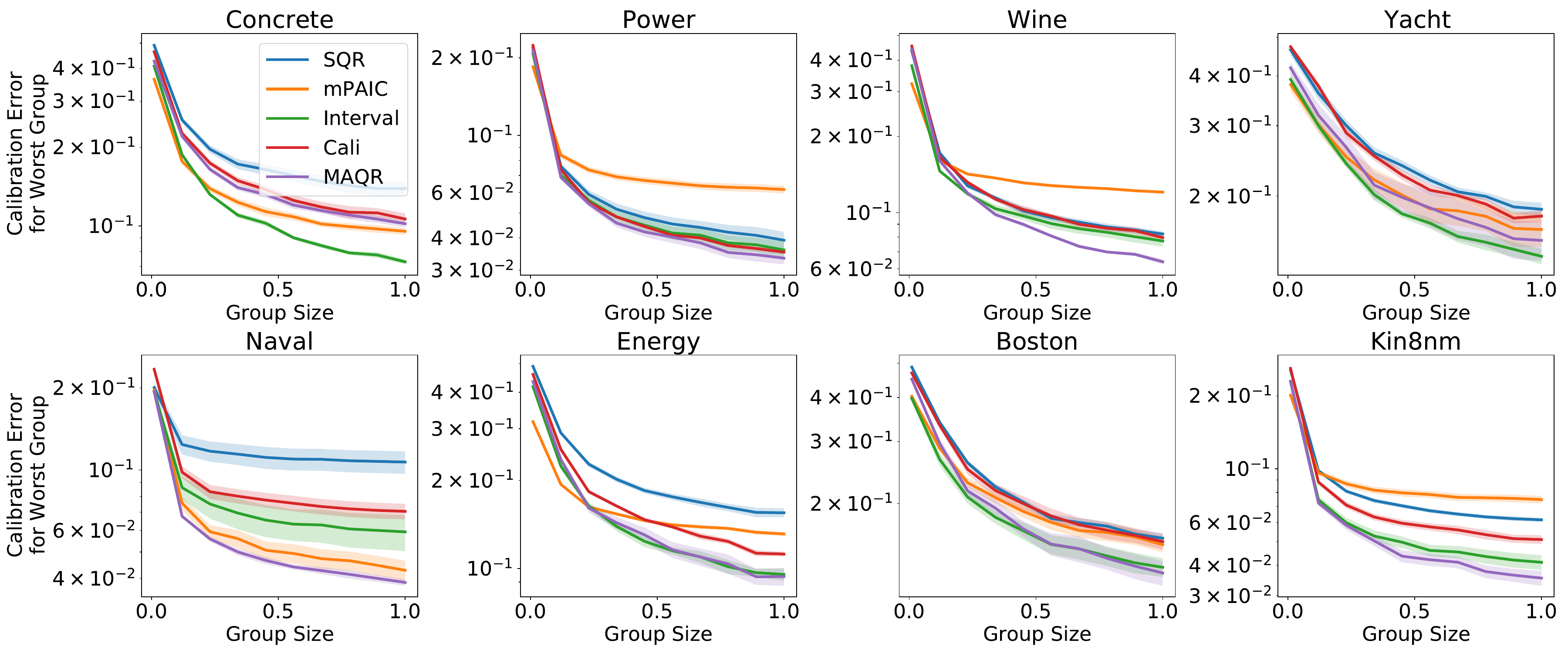}
    \caption{
    \textbf{UCI Adversarial Group Caibration.}
    The plot displays the worst calibration error incurred for any group of any given size. The mean is plotted along with $\pm1$ standard error in shades. Group size here refers to the proportion of the test dataset size
    (full results for Figure~\ref{fig:uci_table_figure} (Bottom)).
    }
    \label{fig:full_uci_gc}
\end{figure}

\begin{figure*}[h!]
\begin{center} \label{table:full_uci_check}
\begin{tabular}{ |c|c|c|c|c|c|c| } 
 \hline
         & \textit{SQR}                 & \textit{mPAIC}            & \textbf{\textit{Interval}}      & \textbf{\textit{Cali}}     & \textbf{\textit{MAQR}}        \\\hline
concrete & $0.085 (0.006)$              & $ 0.085 \pm 0.005 $    & $ 0.086 \pm 0.004 $  & $ 0.118 \pm 0.006$  & $\mathbf{ 0.059 \pm 0.008 }$    \\ \hline
power    & $\mathbf{0.057 \pm 0.001 }$  & $ 0.070 \pm 0.001 $    & $ 0.062 \pm 0.001 $  & $ 0.064 \pm 0.001$  & $0.058 \pm 0.001 $              \\ \hline
wine     & $0.205 \pm 0.008 $           & $ 0.219 \pm 0.004 $    & $ 0.214 \pm 0.006 $  & $ 0.210 \pm 0.008$  & $\mathbf{ 0.191 \pm 0.003}$    \\ \hline
yacht    & $0.012 \pm 0.002 $           & $ 0.015 \pm 0.002 $    & $ 0.018 \pm 0.003 $  & $ 0.019 \pm 0.004$  & $\mathbf{ 0.007 \pm 0.001}$    \\ \hline
naval    & $0.070 \pm 0.001 $           & $ 0.276 \pm 0.004 $    & $ 0.066 \pm 0.013 $  & $ 0.159 \pm 0.029$  & $\mathbf{ 0.004 \pm 0.000}$    \\ \hline
energy   & $0.014 \pm 0.000 $           & $ 0.015 \pm 0.001 $    & $ 0.017 \pm 0.003 $  & $ 0.017 \pm 0.002$  & $\mathbf{ 0.010 \pm 0.001}$    \\ \hline
boston   & $0.088 \pm 0.008 $           & $ 0.095 \pm 0.008 $    & $ 0.094 \pm 0.009 $  & $ 0.103 \pm 0.013$  & $\mathbf{ 0.063 \pm 0.016}$    \\ \hline
kin8nm   & $0.078 \pm 0.001 $           & $ 0.104 \pm 0.003 $    & $ 0.077 \pm 0.001 $  & $ 0.096 \pm 0.005$  & $\mathbf{ 0.070 \pm 0.001}$    \\ \hline
\end{tabular}
\caption{\textbf{UCI Check Score.} Full check score results of UCI experiments from Section~\ref{sec:uci_fusion}.
    Mean score across 5 trials is given, along with $\pm 1$ standard error. 
    The best mean has been bolded.
    \textit{MAQR} tends to achieve the best check score, 
    which is surprising given that \textit{SQR} utilizes the
    same model class to optimize the check score directly.}
\end{center}
\end{figure*}

\begin{figure*}[h!]
\begin{center} \label{table:full_uci_int}
\begin{tabular}{ |c|c|c|c|c|c|c| } 
 \hline
& \textit{SQR}                 & \textit{mPAIC}            & \textbf{\textit{Interval}}      & \textbf{\textit{Cali}}     & \textbf{\textit{MAQR}}        \\\hline
concrete & $ 2.038 \pm 0.225 $   & $ 1.157 \pm 0.069 $  & $ 0.943 \pm 0.053 $  & $ 1.465 \pm 0.086 $  & $ \mathbf{ 0.672 \pm 0.118 }$    \\ \hline
power    & $ 0.834 \pm 0.022 $   & $ 0.917 \pm 0.021 $  & $ 0.620 \pm 0.010 $  & $ 0.699 \pm 0.019 $  & $ \mathbf{0.592 \pm 0.009 }$    \\ \hline
wine     & $ 3.242 \pm 0.166 $   & $ 3.168 \pm 0.019 $  & $ 2.197 \pm 0.045 $  & $ 2.498 \pm 0.135 $  & $ \mathbf{2.052 \pm 0.052 }$    \\ \hline
yacht    & $ 0.314 \pm 0.061 $   & $ 0.197 \pm 0.036 $  & $ 0.190 \pm 0.021 $  & $ 0.298 \pm 0.063 $  & $ \mathbf{0.086 \pm 0.016 }$    \\ \hline
naval    & $ 0.097 \pm 0.011 $   & $ 3.112 \pm 0.053 $  & $ 0.620 \pm 0.114 $  & $ 1.560 \pm 0.268 $  & $ \mathbf{0.044 \pm 0.001 }$    \\ \hline
energy   & $ 0.290 \pm 0.016 $   & $ 0.223 \pm 0.017 $  & $ 0.182 \pm 0.026 $  & $ 0.204 \pm 0.018 $  & $ \mathbf{0.101 \pm 0.006 }$    \\ \hline
boston   & $ 1.833 \pm 0.299 $   & $ 1.395 \pm 0.176 $  & $ 1.010 \pm 0.118 $  & $ 1.449 \pm 0.259 $  & $ \mathbf{0.864 \pm 0.287 }$    \\ \hline
kin8nm   & $ 1.241 \pm 0.041 $   & $ 1.347 \pm 0.031 $  & $ 0.776 \pm 0.017 $  & $ 1.121 \pm 0.072 $  & $ \mathbf{0.691 \pm 0.015 }$    \\ \hline
\end{tabular}
\caption{\textbf{UCI Interval Score} Full interval score results 
of UCI experiments from Section~\ref{sec:uci_fusion}.
    Mean score across 5 trials is given, along with $\pm 1$ standard error. 
    The best mean has been bolded.
    \textit{MAQR} tends to achieve the best interval score, 
    which is surprising given that \textit{Interval} utilizes the
    same model class to optimize the interval score directly.}
\end{center}
\end{figure*}

\begin{figure*}[h!]
\begin{center} \label{table:uci_int_cali}
\begin{tabular}{ |c|c|c|c|c|c|c| } 
 \hline
& \textit{SQR}                 & \textit{mPAIC}            & \textbf{\textit{Interval}}      & \textbf{\textit{Cali}}     & \textbf{\textit{MAQR}}        \\\hline
concrete & $ 0.186 \pm 0.031 $   & $ 0.089 \pm 0.005 $  & $ 0.061 \pm 0.008 $  & $ 0.096 \pm 0.013 $  & $ \mathbf{0.059 \pm 0.020 }$    \\ \hline
power    & $ 0.045 \pm 0.004 $   & $ 0.068 \pm 0.008 $  & $ 0.023 \pm 0.003 $  & $ 0.037 \pm 0.002 $  & $ \mathbf{0.010 \pm 0.002 }$    \\ \hline
wine     & $ 0.053 \pm 0.006 $   & $ 0.169 \pm 0.008 $  & $ 0.079 \pm 0.014 $  & $ 0.065 \pm 0.007 $  & $ \mathbf{0.045 \pm 0.005 }$    \\ \hline
yacht    & $ 0.135 \pm 0.009 $   & $ 0.100 \pm 0.020 $  & $ 0.121 \pm 0.005 $  & $ 0.129 \pm 0.016 $  & $ \mathbf{0.085 \pm 0.024 }$    \\ \hline
naval    & $ 0.128 \pm 0.031 $   & $ 0.039 \pm 0.003 $  & $ 0.043 \pm 0.014 $  & $ 0.110 \pm 0.013 $  & $ \mathbf{0.012 \pm 0.002 }$    \\ \hline
energy   & $ 0.174 \pm 0.011 $   & $ 0.163 \pm 0.009 $  & $ 0.060 \pm 0.010 $  & $ 0.090 \pm 0.011 $  & $ \mathbf{0.052 \pm 0.018 }$    \\ \hline
boston   & $ 0.163 \pm 0.020 $   & $ \mathbf{0.050 \pm 0.007 } $  & $ 0.079 \pm 0.015 $  & $ 0.138 \pm 0.028 $  & $ 0.092 \pm 0.041 $    \\ \hline
kin8nm   & $ 0.070 \pm 0.005 $   & $ 0.080 \pm 0.002 $  & $ 0.048 \pm 0.006 $  & $ 0.067 \pm 0.005 $  & $ \mathbf{0.019 \pm 0.008 }$    \\ \hline
\end{tabular}
\caption{\textbf{UCI Centered Interval Calibration} Full centered interval calibration results 
of UCI experiments from Section~\ref{sec:uci_fusion}.
    Mean score across 5 trials is given, along with $\pm 1$ standard error. 
    The best mean has been bolded.
    \textit{MAQR} tends to achieve the best centered interval calibration. \label{table:app_uci_centered_interval}}
\end{center}
\end{figure*}

\clearpage
\subsection{Full Fusion Experiment Results} \label{app:all_fusion_results}

\begin{figure*}[h!]
    \centering
    \includegraphics[width=\textwidth]{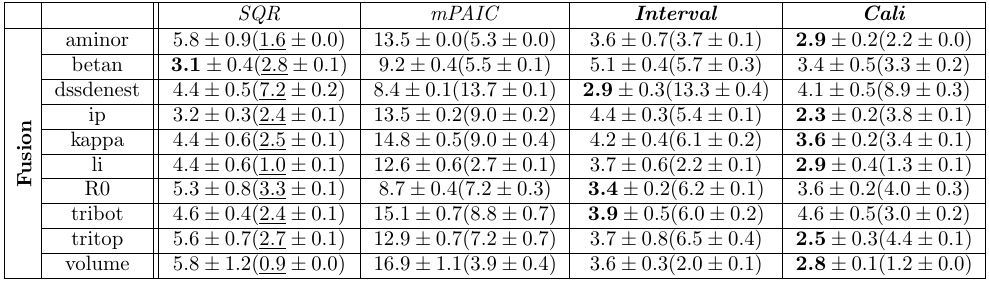}
    \caption{\textbf{Fusion Average Calibration-Sharpness Table.} The table shows mean average calibration (measured by ECE) and sharpness in parentheses, along with $\pm 1$ standard error. 
    The best mean ECE for each dataset has been bolded and the best mean sharpness has been underlined. 
    All values have been multiplied by 100 for readability.
    \textit{Cali} tends to achieve the best average calibration,
    while \textit{SQR} achieves the sharpest predictions (same table as Figure~\ref{fig:fusion_table_figure} (Top), repeated here for completeness)
    }
    \label{fig:fusion_ece_sharp_table_full}
\end{figure*}

\begin{figure*}[h!]
    \centering
    \includegraphics[width=\textwidth]{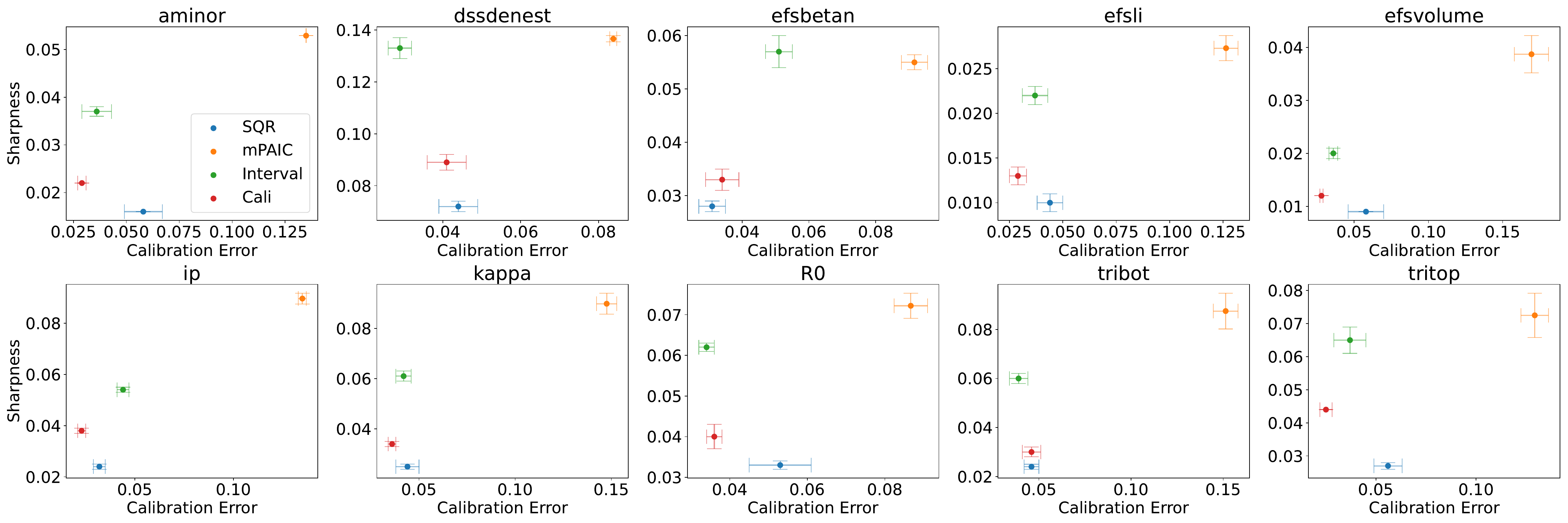}
    \caption{\textbf{Fusion Average Calibration-Sharpness Plot.} Visualization of average calibration-sharpness from fusion experiments from Section~\ref{sec:uci_fusion}}.
    \label{fig:full_fusion_cal_sharp}
\end{figure*}

\begin{figure*}[h!]
    \centering
    \includegraphics[width=\textwidth]{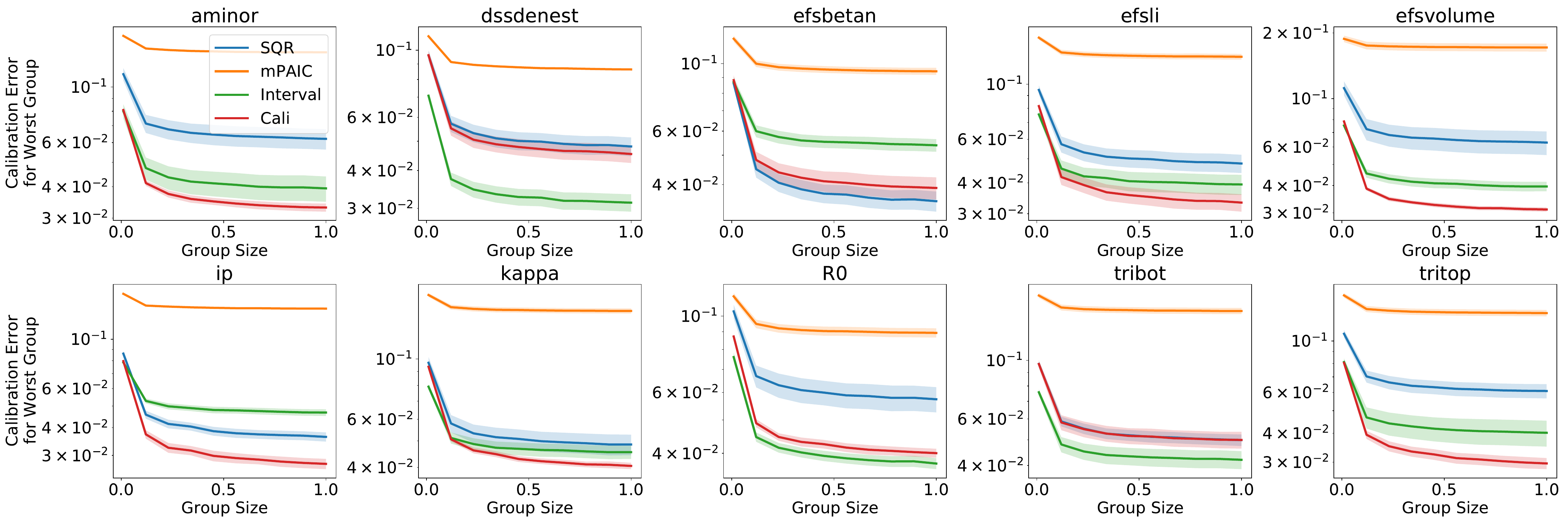}
    \caption{\textbf{Fusion Adversarial Group Calibration}  Full fusion results of Figure~\ref{fig:fusion_table_figure} (Bottom).
    \textit{Cali} and \textit{Interval} tend to achieve the best
    calibration for any group of any size in the test set.}
    \label{fig:full_fusion_gc}
\end{figure*}

\begin{figure*}[h!]
\begin{center} \label{table:full_fusion_check}
\begin{tabular}{ |c|c|c|c|c|c|c| } 
 \hline
 & \textit{SQR}                 & \textit{mPAIC}            & \textbf{\textit{Interval}}               & \textbf{\textit{Cali}}                      \\\hline
aminor     & $ \mathbf{0.087 \pm 0.000 } $   & $ 0.182 \pm 0.000 $  & $ 0.097 \pm 0.002 $  & $ 0.092 \pm 0.002 $ \\ \hline
dssdenest  & $ \mathbf{0.179 \pm 0.003 } $   & $ 0.273 \pm 0.003 $  & $ 0.180 \pm 0.001 $  & $ 0.184 \pm 0.002 $ \\ \hline
betan      & $ \mathbf{0.146 \pm 0.001 } $   & $ 0.253 \pm 0.005 $  & $ 0.153 \pm 0.002 $  & $ 0.150 \pm 0.001 $ \\ \hline
li         & $ \mathbf{0.097 \pm 0.001 } $   & $ 0.166 \pm 0.003 $  & $ 0.102 \pm 0.001 $  & $ 0.104 \pm 0.001 $ \\ \hline
volume     & $ \mathbf{0.051 \pm 0.001 } $   & $ 0.107 \pm 0.007 $  & $ 0.053 \pm 0.001 $  & $ 0.052 \pm 0.001 $ \\ \hline
ip         & $ \mathbf{0.068 \pm 0.000 } $   & $ 0.199 \pm 0.011 $  & $ 0.077 \pm 0.002 $  & $ 0.076 \pm 0.001 $ \\ \hline
kappa      & $ \mathbf{0.072 \pm 0.001 } $   & $ 0.150 \pm 0.004 $  & $ 0.079 \pm 0.002 $  & $ 0.078 \pm 0.001 $ \\ \hline
R0         & $ \mathbf{0.120 \pm 0.001 } $   & $ 0.208 \pm 0.002 $  & $ 0.126 \pm 0.002 $  & $ 0.130 \pm 0.005 $ \\ \hline
tribot     & $ \mathbf{0.084 \pm 0.001 } $   & $ 0.184 \pm 0.020 $  & $ 0.092 \pm 0.001 $  & $ 0.096 \pm 0.005 $ \\ \hline
tritop     & $ \mathbf{0.102 \pm 0.001 } $   & $ 0.200 \pm 0.018 $  & $ 0.107 \pm 0.004 $  & $ 0.107 \pm 0.002 $ \\ \hline
\end{tabular}
\caption{\textbf{Fusion Check Score} Check score results from fusion experiments in Section~\ref{sec:uci_fusion}.
    Mean score across 5 trials is given, along with $\pm 1$ standard error. 
    The best mean has been bolded.
    \textit{SQR} achieves the best check score.}
\end{center}
\end{figure*}

\begin{figure*}[h!]
\begin{center} \label{table:fusion_int}
\begin{tabular}{ |c|c|c|c|c|c|c| } 
 \hline
& \textit{SQR}                 & \textit{mPAIC}            & \textbf{\textit{Interval}}               & \textbf{\textit{Cali}}                      \\\hline
aminor     & $ 1.181 \pm 0.007 $   & $ 3.225 \pm 0.000 $  & $ \mathbf{1.090 \pm 0.017 }$  & $ 1.207 \pm 0.035$ \\ \hline
dssdenest  & $ 2.387 \pm 0.060 $   & $ 4.352 \pm 0.109 $  & $ \mathbf{1.995 \pm 0.011 }$  & $ 2.369 \pm 0.054$ \\ \hline
betan      & $ 1.970 \pm 0.013 $   & $ 4.301 \pm 0.124 $  & $ \mathbf{1.725 \pm 0.021 }$  & $ 2.002 \pm 0.046$ \\ \hline
li         & $ 1.354 \pm 0.018 $   & $ 2.923 \pm 0.088 $  & $ \mathbf{1.146 \pm 0.010 }$  & $ 1.410 \pm 0.015$ \\ \hline
volume     & $ 0.711 \pm 0.023 $   & $ 2.036 \pm 0.154 $  & $ \mathbf{0.602 \pm 0.012 }$  & $ 0.697 \pm 0.026$ \\ \hline
ip         & $ 0.906 \pm 0.005 $   & $ 3.203 \pm 0.295 $  & $ \mathbf{0.845 \pm 0.017 }$  & $ 0.978 \pm 0.026$ \\ \hline
kappa      & $ 1.010 \pm 0.010 $   & $ 2.639 \pm 0.101 $  & $ \mathbf{0.891 \pm 0.015 }$  & $ 1.066 \pm 0.020$ \\ \hline
R0         & $ 1.599 \pm 0.011 $   & $ 3.310 \pm 0.064 $  & $ \mathbf{1.414 \pm 0.012 }$  & $ 1.709 \pm 0.057$ \\ \hline
tribot     & $ 1.211 \pm 0.009 $   & $ 3.185 \pm 0.242 $  & $ \mathbf{1.074 \pm 0.012 }$  & $ 1.421 \pm 0.115$ \\ \hline
tritop     & $ 1.481 \pm 0.022 $   & $ 3.564 \pm 0.351 $  & $ \mathbf{1.232 \pm 0.034 }$  & $ 1.444 \pm 0.031$ \\ \hline
\end{tabular}
\caption{\textbf{Fusion Interval Score} 
    Interval score results from fusion experiments in  Section~\ref{sec:uci_fusion}.
    Mean score across 5 trials is given, along with $\pm 1$ standard error. 
    The best mean has been bolded.
    \textit{Interval} achieves the best interval score.}
\end{center}
\end{figure*}

\begin{figure*}[h!]
\begin{center} \label{table:fusion_int_cali}
\begin{tabular}{ |c|c|c|c|c|c|c| } 
 \hline
& \textit{SQR}                 & \textit{mPAIC}            & \textbf{\textit{Interval}}               & \textbf{\textit{Cali}}                      \\\hline
aminor     & $ 0.081 \pm 0.006 $   & $ 0.268 \pm 0.000 $  & $\mathbf{ 0.055 \pm 0.007 } $ & $ 0.058 \pm 0.004 $ \\ \hline
dssdenest  & $ 0.072 \pm 0.003 $   & $ 0.162 \pm 0.003 $  & $\mathbf{ 0.054 \pm 0.006 } $ & $ 0.071 \pm 0.008 $ \\ \hline
betan      & $\mathbf{0.051 \pm 0.001 }$   & $ 0.160 \pm 0.012  $  & $ 0.087 \pm 0.009 $  & $ 0.056 \pm 0.007 $ \\ \hline
li         & $ 0.071 \pm 0.011 $   & $ 0.214 \pm 0.017 $  & $ 0.069 \pm 0.013 $  & $\mathbf{ 0.043 \pm 0.003 } $\\ \hline
volume     & $ 0.073 \pm 0.007 $   & $ 0.315 \pm 0.022 $  & $ 0.063 \pm 0.007 $  & $\mathbf{ 0.056 \pm 0.002 } $\\ \hline
ip         & $ 0.051 \pm 0.003 $   & $ 0.227 \pm 0.020 $  & $ 0.077 \pm 0.015 $  & $\mathbf{ 0.036 \pm 0.003 } $\\ \hline
kappa      & $ 0.069 \pm 0.008 $   & $ 0.254 \pm 0.014 $  & $ 0.078 \pm 0.011 $  & $\mathbf{ 0.053 \pm 0.004 } $\\ \hline
R0         & $ 0.068 \pm 0.007 $   & $ 0.168 \pm 0.009 $  & $\mathbf{ 0.058 \pm 0.007 } $ & $ 0.066 \pm 0.004 $ \\ \hline
tribot     & $ 0.087 \pm 0.006 $   & $ 0.227 \pm 0.016 $  & $\mathbf{ 0.074 \pm 0.008 } $ & $ 0.080 \pm 0.007 $ \\ \hline
tritop     & $ 0.077 \pm 0.006 $   & $ 0.231 \pm 0.017 $  & $ 0.058 \pm 0.015 $  & $ \mathbf{0.037 \pm 0.004 }$ \\ \hline
\end{tabular}
\caption{\textbf{Fusion Centered Interval Calibration}
     Full centered interval calibration results from fusion experiments in Section~\ref{sec:uci_fusion}.
    Mean score across 5 trials is given, along with $\pm 1$ standard error. 
    The best mean has been bolded.
    \textit{Cali} and \textit{Interval} achieves the best centered
    interval calibration in 9 out of 10 datasets.}
\end{center}
\end{figure*}

%% file: app_add_exp.tex
\section{Additional Experiments}
\subsection{Details of Synthetic Experiment (Figure~\ref{fig:synth_example})}
\label{app:synth_example}
\textbf{Dataset}: The synthetic dataset is based on the Boston
UCI dataset. 
A NN model, $\mu_\theta$, was trained on the train split of the Boston dataset to predict the mean by optimizing the MSE loss (same architecture and training details as described in Appendix~\ref{app:uci_exp_details}).
Afterwards, all points, $\{(x_i, y_i)\}_{i=1}^{N}$, in the full Boston dataset were re-labelled with the prediction of the mean model, $(x_i, \mu_\theta(x_i))$. 
Then, uniform noise, $\eps_i$ was added to these re-labelled mean values to create the observations, $\Tilde{y}_i$, i.e. $\Tilde{y}_i = \mu_\theta(x_i) + \eps_i$. 
The uniform noise was 0 mean, with width of the support equal to $5\%$ of the range of the mean values, i.e $\eps_i \sim$ Unif$[-0.025*(\max_i{y_i} - \min_i{y_i}), 0.025*(\max_i{y_i} - \min_i{y_i})]$.
Thus synthetic dataset is $\{(x_i, \Tilde{y_i})\}_{i=1}^{N}$.

\textbf{Procedure}: The training procedure follows exactly that of the main experiments, which is described in detail in Appendix~\ref{app:uci_exp_details}.
The only difference is that the models were trained for the full 1000 epochs, instead of halting training according to the validation loss.

\subsection{Regularizing the Pinball Loss}
\label{app:regularization_pinball}
At first glance, regularization may appear to be the answer to what seems like 
an overfitting problem with the pinball loss.
In fact, regularization has shown to be effective in the ``single quantile learning setting'', where the quantile level $p$ is fixed and the $p\textsuperscript{th}$ conditional quantile is learned from data \citep{steinwart2011estimating, takeuchi2006nonparametric}.
This setting is concerned with learning $\Q(x)$ for a \textit{given} $p$, 
which is different from the setting of this work, 
which considers learning a single model $\Q_p(x)$, 
which takes both $x$ and $p$ as input and outputs the full predictive distribution by specifying conditional quantiles predictions for all quantiles levels.

In this section, we empirically demonstrate the effect of applying regularization when minimizing the pinball loss to learn $\Q_p(x)$, and show how regularization does not 
adequately address the issues with the pinball loss.

With the \textit{SQR} method (which optimizes the pinball loss simultaneously for random batches of quantile levels),
we applied L1, L2, and dropout, 
by cross-validating the regularization coefficients 
in \{$2^{i}, i \in $ np.linspace$(-13, 1, 40)$\} for L1 and L2, 
and dropout probability $p$ 
in \{$2^{i}, i \in $ np.linspace$(-13, -1, 40)$\},
for the pinball loss criterion (i.e. cross-validate among 40 different regularization coefficients, between $2^{-13}$ and $2^{-1}$ on the log scale).  
We show the best calibrated regularization result, across all regularization methods and cross-validation (table in same format as Figure 2 in paper).
\begin{figure*}[h!]
\centering
\begin{tabular}{ |c|c|c|c|c|c|c| } 
 \hline
          & \textit{SQR} & \textit{SQR w/Reg}      \\ \hline
concrete & $ \mathbf{9.3} \pm 1.5 (7.0 \pm 1.0)$ & $ 10.9 \pm 1.0 (\underline{6.9} \pm 0.6)$  \\ \hline
power    & $ 2.6 \pm 0.4 (\underline{13.4} \pm 0.2)$ & $ \mathbf{1.0} \pm 0.1 (14.8 \pm 0.1)$  \\ \hline
wine     & $ \mathbf{4.2} \pm 0.2 (29.5 \pm 0.4)$ & $ 5.1 \pm 0.8 (\underline{26.0} \pm 0.5)$  \\ \hline
yacht    & $ \mathbf{9.4} \pm 0.9 (\underline{1.0} \pm 0.1)$ & $ 12.3 \pm 2.6 (\underline{1.0} \pm 0.1)$   \\ \hline
naval    & $ \mathbf{9.7} \pm 1.6 (\underline{3.5} \pm 0.4) $ & $ 11.5 \pm 2.8 (\underline{3.5} \pm 0.3)$  \\ \hline
energy   & $ 9.8 \pm 0.8 (2.0 \pm 0.1) $ & $ \mathbf{9.4} \pm 1.3 (\underline{1.9} \pm 0.2)$   \\ \hline
boston   & $ \mathbf{9.0} \pm 0.8 (9.3 \pm 0.7) $ & $ 11.6 \pm 1.1 (\underline{8.6} \pm 0.8)$  \\ \hline
kin8nm   & $ 4.4 \pm 0.1 (11.4 \pm 0.2)$ & $ \mathbf{3.5} \pm 0.3 (\underline{11.2} \pm 0.2)$  \\ \hline
\end{tabular}
\caption{\textbf{Applying Regularization with SQR: Average Calibration and Sharpness.}
The table shows \textit{SQR}'s mean ECE and sharpness (in parentheses) and their standard error with and without regularization. Among the 3 regularization methods (L1, L2, dropout), the method resulting in the best calibration is shown, for each dataset. The best mean ECE for each dataset has been bolded and the best mean sharpness has been underlined. All values have been multiplied by 100 for readability.}
\end{figure*}

Counter-intuitively, regularization tends to further the bias towards sharpness,
and upon reflection, this may not be surprising because the range of quantile
predictions shrinks: given a quantile model $f: \mathcal{X} \times \mathcal{P} \rightarrow \mathcal{Y}$, where $\mathcal{P}$ is the space of quantile levels in $(0, 1)$, regularization affects the smoothness not only in $\mathcal{X}$, but also in $\mathcal{P}$, hence for any fixed $x\in\mathcal{X}$, the range in predictions for different quantile level inputs also shrinks.

\subsection{Comparison on 95\% Prediction Interval Task} \label{app:coverage_95}
In Section~\ref{sec:experiments}, we have presented an experiment that is targeted at evaluating the \textit{full predictive distribution} on the 8 UCI datasets.

In this section, we present an experiment that is targeted at constructing a \textit{$95\%$ centered prediction interval (PI)} on the same UCI datasets, for the purposes of comparing against other quantile-based algorithms that are designed to output only
a single quantile level.
This experiment setup is exactly the same as the prediction intervals experiments 
in Section 4.1 of the work by \citet{tagasovska2019single}, and we follow the exact same experiment procedure 
for a direct comparison against their reported results.

The comparison algorithms here are:
\begin{itemize}
    \item Dropout \citep{gal2016dropout}:
    a NN that uses a dropout layer during testing for multiple predictions
    \item QualityDriven \citep{pearce2018high}:
    a NN that optimizes a Binomial likelihood approximation as a surrogate loss for calibration and sharpness 
    \item GradientBoosting \citep{meinshausen2006quantile}:
    a decision tree based model that optimizes the pinball loss with gradient boosting
    \item QuantileForest \citep{meinshausen2006quantile}:
    a decision tree based model that predicts the quantiles based on the trained 
    output of a random forest
    \item ConditionalGaussian \citep{lakshminarayanan2017simple}: 
    a probabilistic NN that optimizes the Gaussian NLL to output the parameters 
    of a Gaussian distribution 
\end{itemize}

We show the performance of \textit{MAQR} to represent our proposed methods in this experiment, 
since \textit{MAQR} performs the best on the full predictive distribution evaluations in the UCI experiments of Section~\ref{sec:uci_fusion}.
We also omit the results of \textit{SQR} \citet{tagasovska2019single} in this experiment since we perform
a full evaluation comparison with \textit{SQR} in Section~\ref{sec:uci_fusion} and Appendix~\ref{app:all_uci_results},  \ref{app:all_fusion_results}, 
and the purpose here is to compare our proposed method against the additional baselines.

In this experiment, since we only output two quantile levels ($0.025, 0.975$) to 
construct a single $95\%$ prediction interval, we do not assess calibration (which requires the predictions for all quantile levels)
and assess only the observed proportion of test points within the PI (also referred to as ``prediction interval coverage probability'' or ``PICP'') and sharpness 
represented by the width of the PI (also referred to as ``mean prediction interval width'' or ``MPIW'').
We refer the reader to \citet{tagasovska2019single} for exact details on the experiment setup and the hyperparameters tuned.
For \textit{MAQR}, we used the exact same NN architecture (1 layer of 64 hidden units, ReLU non-linearities) as the NN based baselines (\textit{Dropout, QualityDriven, ConditionalGaussian}) and the same training procedure as detailed in \citet{tagasovska2019single}.
The hyperparameters tuned for \textit{MAQR} are detailed in Appendix~\ref{app:uci_exp_details}.

\begin{figure*}[h!] 
\begin{center}
\begin{tabular}{ |c|c|c|c|c|c|c| } 
\hline
            & \textit{Dropout}                        & \textit{QualityDriven}                  & \textit{GradientBoostingQR}                \\ \hline       
 concrete   & none                           & none                           & $0.93\pm0.00$ $(0.71\pm0.00)$     \\ \hline
 power      & $0.94\pm0.00$ $(0.37\pm0.00)$  & $0.93\pm0.02$ $(0.34\pm0.19)$  & none                              \\ \hline
 wine       & none                           & none                           & none                              \\ \hline
 yacht      & $0.97\pm0.03$ $(0.10\pm0.01)$  & $0.92\pm0.05$ $(0.04\pm0.01)$  & $0.95\pm0.02$ $(0.79\pm0.01)$     \\ \hline
 naval      & $0.96\pm0.01$ $(0.23\pm0.00)$  & $0.94\pm0.02$ $(0.21\pm0.11)$  & none                              \\ \hline
 energy     & $0.91\pm0.04$ $(0.17\pm0.01)$  & $0.91\pm0.04$ $(0.10\pm0.05)$  & none                              \\ \hline
 boston     & none                           & none                           & $0.89\pm0.00$ $(0.75\pm0.00)$     \\ \hline
 kin8nm     & none                           & $0.96\pm0.00$ $(0.84\pm0.00)$  & none                              \\ \hline    
\hline
            & \textit{QuantileForest}                 & \textit{ConditionalGaussian}            & \textbf{\textit{MAQR}}            \\ \hline
 concrete   & $0.96\pm0.01$ $(0.37\pm0.02)$  & $0.94\pm0.03$ $(0.32\pm0.09)$  & $0.93\pm0.01$ $(0.26\pm0.01)$     \\ \hline
 power      & $0.94\pm0.01$ $(0.18\pm0.00)$  & $0.94\pm0.01$ $(0.18\pm0.00)$  & $0.95\pm0.01$ $(0.28\pm0.03)$     \\ \hline
 wine       & none                           & $0.94\pm0.02$ $(0.49\pm0.03)$  & $0.95\pm0.02$ $(0.56\pm0.06)$     \\ \hline
 yacht      & $0.97\pm0.04$ $(0.28\pm0.11)$  & $0.93\pm0.06$ $(0.03\pm0.01)$  & $0.92\pm0.03$ $(0.03\pm0.01)$     \\ \hline
 naval      & $0.92\pm0.01$ $(0.22\pm0.00)$  & $0.96\pm0.01$ $(0.15\pm0.25)$  & $0.94\pm0.01$ $(0.03\pm0.00)$     \\ \hline
 energy     & $0.95\pm0.02$ $(0.15\pm0.01)$  & $0.94\pm0.03$ $(0.12\pm0.18)$  & $0.94\pm0.02$ $(0.05\pm0.01)$     \\ \hline
 boston     & $0.95\pm0.03$ $(0.37\pm0.02)$  & $0.94\pm0.03$ $(0.55\pm0.20)$  & $0.95\pm0.02$ $(0.34\pm0.09)$     \\ \hline
 kin8nm     & none                           & $0.93\pm0.01$ $(0.20\pm0.01)$  & $0.93\pm0.00$ $(0.20\pm0.01)$     \\ \hline
\hline

\end{tabular}
\caption{\textbf{95\% PI PICP and MPIW} The test average and standard deviation PICP of models with validation PICP in $[92.5\%, 97.5\%]$ is shown, and the test average and standard deviation MPIW is shown in parantheses. ``none'' indicates the method could not find a model with validation PICP in $[92.5\%, 97.5\%]$.}\label{tab:pi_95}
\end{center}
\end{figure*}

Summarizing the experimental result in Figure~\ref{tab:pi_95}:
\begin{itemize}
    \item 
    Our proposed method (\textit{MAQR}) is capable of consistently producing PIs
    that have the correct desired coverage ($0.95$),  
    even in cases when some of the baseline algorithms are not able to.
    \item
    Even when the baseline algorithms do produce PIs with correct desired coverage, 
    \textit{MAQR} is able to produce PI's that are much sharper (e.g. mean PI width for naval dataset
    is an order of magnitude sharper than all other baselines) 
\end{itemize}

On this limited output and evaluation setting, our proposed method is still competitive
in its performance.
However, it should be noted that this experiment tells \textit{only one facet} of overall UQ quality.
Inspecting and evaluating the full predictive uncertainty is necessary for a more thorough evaluation of UQ quality, as done in our main experiments in Section~\ref{sec:experiments}.

\subsection{Discussion on Recalibration \citep{kuleshov2018accurate}} \label{app:recal}
The recalibration algorithm by \citet{kuleshov2018accurate} utilizes isotonic regression with a validation set to fine-tune predictive uncertainties from a UQ model.
We have applied this recalibration as a post-processing step on the methods presented in Section~\ref{sec:experiments}, 
and the empirical results indicate that its effect on overall improvement in UQ quality is inconclusive.
Here, we show its effect on one of our methods, \textit{Interval}, because we observe the same pattern across all the methods, including the baseline algorithms.

Recalibration tends to improve average calibration.
This is expected, because recalibration specifically targets average calibration.
However, it does so at a cost in sharpness. This is evident in the recalibrated output moving 
upper left in the average calibration-sharpness plot in Figure~\ref{fig:uci_recal_cal_sharp}.

However, there is little to no improvement in adversarial group calibration (except for the Naval dataset) as shown in Figure~\ref{fig:uci_recal_gc}, which seems to indicate that the improvement in average calibration was not meaningful (i.e. the recalibrated result is not closer to individual calibration).
This is also the observation made by \citet{zhao2020individual}.

At the same time, the proper scoring rules improved on average (Figures~\ref{table:uci_recal_check}, \ref{table:uci_recal_int}), but interval calibration tended to worsen (Figure~\ref{table:uci_recal_int_cali}).

Based on these metrics, it is difficult to conclude on whether recalibration by \citet{kuleshov2018accurate} is 
a beneficial step or not for overall UQ quality.
If a practitioner is primarily concerned with average calibration,
the results indicate that recalibration \textit{is} a beneficial step, 
but if converging to the true conditional distribution is the primary objective, recalibration does not seem to be a robust remedy.

\begin{figure*}[h!] 
\begin{center} 
\begin{tabular}{ |c|c|c|c|c|c|c| } 
 \hline
         & \textbf{\textit{Interval}}                   & \textbf{\textit{Interval Recalibrated}}     \\ \hline
concrete & $ 3.7 \pm 0.6	(\underline{18.1} \pm 0.6)$	& $\mathbf{3.1} \pm 0.4	(26.3 \pm 1.8)$ \\ \hline
power    & $ 2.2 \pm 0.4	(21.0 \pm 1.0)$	& $\mathbf{1.0} \pm 0.1	(\underline{20.6} \pm 0.5)$ \\ \hline
wine     & $ 5.0 \pm 0.8	(\underline{41.4} \pm 2.5)$	& $\mathbf{2.6} \pm 0.2	(50.5 \pm 1.8)$ \\ \hline
yacht    & $ 7.5 \pm 0.9	( \underline{4.5} \pm 1.0)$	& $\mathbf{4.7} \pm 0.5	( 5.6 \pm 1.1)$ \\ \hline
naval    & $ 4.7 \pm 1.4	(28.4 \pm 3.6)$	& $\mathbf{1.3} \pm 0.1	(\underline{21.9} \pm 3.3)$ \\ \hline
energy   & $ 4.3 \pm 0.6	( \underline{5.1} \pm 0.9)$	& $\mathbf{3.8} \pm 0.6	( 6.9 \pm 1.2)$ \\ \hline
boston   & $ 6.9 \pm 1.1	(\underline{20.3} \pm 0.5)$	& $\mathbf{5.4} \pm 0.9	(30.8 \pm 2.6)$ \\ \hline
kin8nm   & $ 2.9 \pm 0.4	(\underline{16.9} \pm 0.5)$	& $\mathbf{1.1} \pm 0.1	(20.6 \pm 0.3)$ \\ \hline
\end{tabular}
\caption{\textbf{UCI Average Calibration-Sharpness Table with Recalibration.} Recalibration
    tends to trade off sharpness for average calibration. 
    Better mean average calibration has been bolded, and better sharpness has been underlined.
    All values have been multiplied by 100 for readability.}
\label{table:uci_recal_cal_sharp_table}
\end{center}
\end{figure*}

\begin{figure*}[h!]
    \centering
    \includegraphics[width=\textwidth]{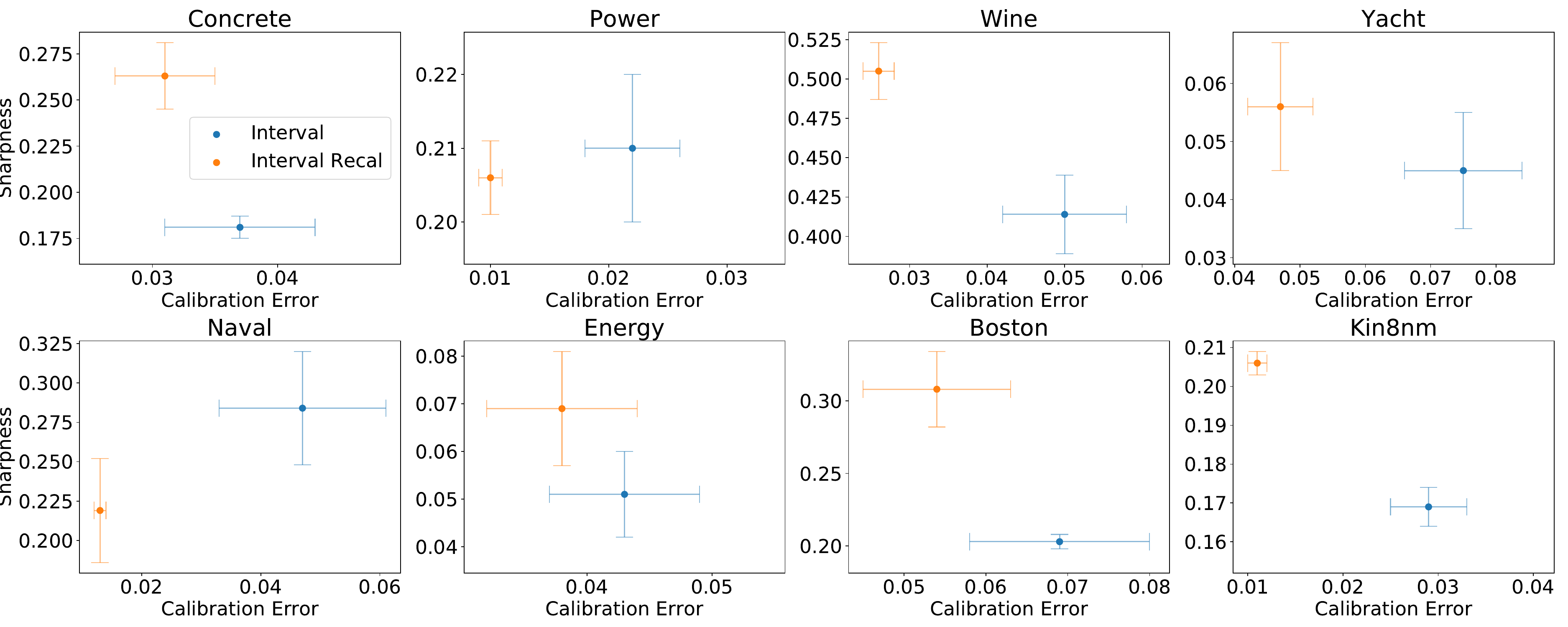}
    \caption{\textbf{UCI Average Calibration-Sharpness Plots with Recalibration.} Recalibration
    tends to trade off sharpness for average calibration. This is evident as the recalibrated
    predictions move \textit{upper left}.}
    \label{fig:uci_recal_cal_sharp}
\end{figure*}

\begin{figure*}[h!]
    \centering
    \includegraphics[width=\textwidth]{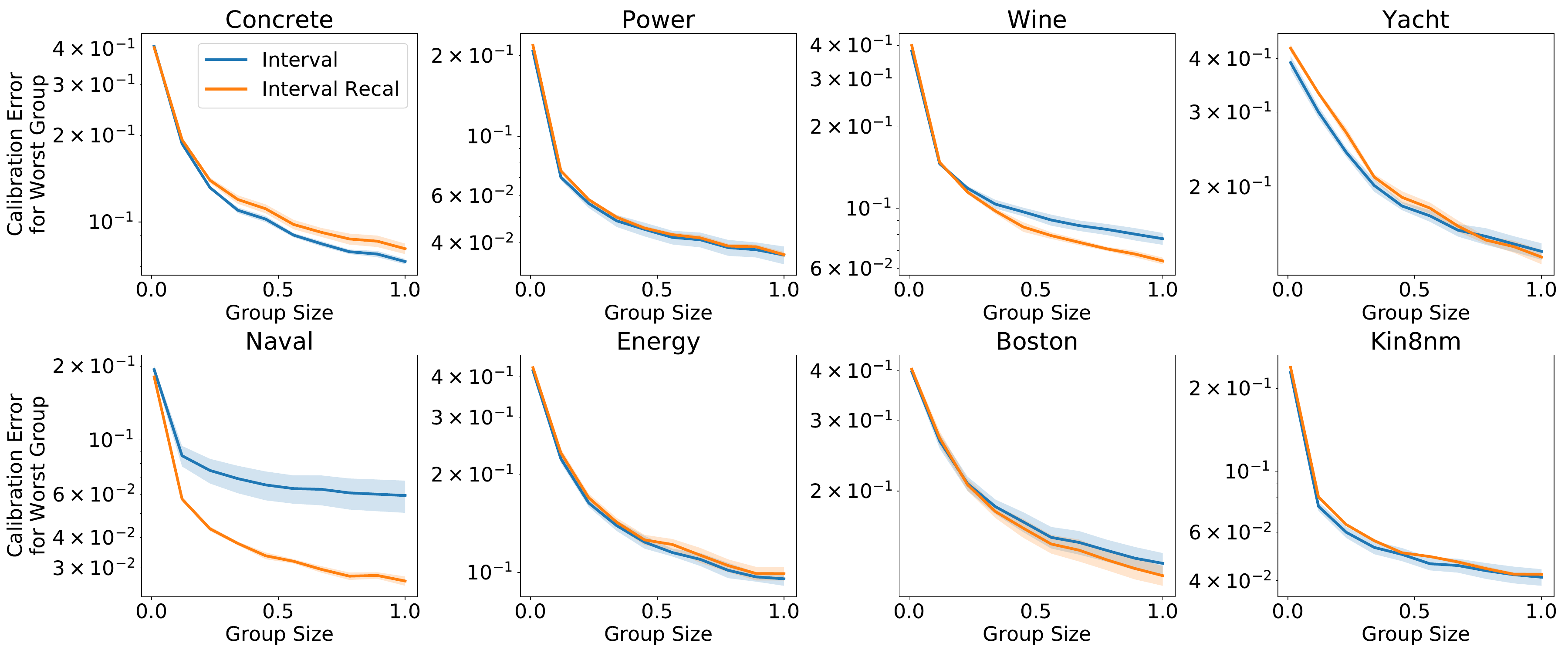}
    \caption{\textbf{UCI Adversarial Group Calibration with Recalibration.} Recalibration in general shows little improvement in adversarial group calibration.}
    \label{fig:uci_recal_gc}
\end{figure*}

\begin{figure*}[h!] 
\begin{center} 
\begin{tabular}{ |c|c|c|c|c|c|c| } 
 \hline
         & \textbf{\textit{Interval}}                   & \textbf{\textit{Interval Recalibrated}}     \\ \hline
concrete & $ 0.086 \pm 0.004 $           & $ \mathbf{0.077 \pm 0.004 }$  \\ \hline
power    & $ 0.062 \pm 0.001 $           & $ \mathbf{0.061 \pm 0.001 }$  \\ \hline
wine     & $ 0.214 \pm 0.006 $           & $ \mathbf{0.209 \pm 0.013 }$  \\ \hline
yacht    & $ \mathbf{0.018 \pm 0.003 }$  & $ \mathbf{0.018 \pm 0.004 }$  \\ \hline
naval    & $ 0.066 \pm 0.013 $           & $ \mathbf{0.062 \pm 0.012 }$  \\ \hline
energy   & $ 0.017 \pm 0.003 $           & $ \mathbf{0.016 \pm 0.003 }$  \\ \hline
boston   & $ 0.094 \pm 0.009 $           & $ \mathbf{0.076 \pm 0.014 }$  \\ \hline
kin8nm   & $ \mathbf{0.077 \pm 0.001 }$  & $ \mathbf{0.077 \pm 0.002 }$  \\ \hline
\end{tabular}
\caption{\textbf{UCI Check Score with Recalibration.} Recalibration tends to improve the check score.}
\label{table:uci_recal_check}
\end{center}
\end{figure*}

\begin{figure*}[h!]
\begin{center} 
\begin{tabular}{ |c|c|c|c|c|c|c| } 
 \hline
          & \textbf{\textit{Interval}}                    & \textbf{\textit{Interval Recalibrated}}       \\ \hline
concrete & $ 0.943 \pm 0.053 $           & $ \mathbf{0.778 \pm 0.064}$   \\ \hline
power    & $ 0.620 \pm 0.010 $           & $ \mathbf{0.616 \pm 0.006}$   \\ \hline
wine     & $ 2.197 \pm 0.045 $           & $ \mathbf{1.921 \pm 0.024}$   \\ \hline
yacht    & $ 0.190 \pm 0.021 $           & $ \mathbf{0.158 \pm 0.025}$   \\ \hline
naval    & $ \mathbf{3.112 \pm 0.053} $  & $ 3.150 \pm 0.050$   \\ \hline
energy   & $ 0.182 \pm 0.026 $           & $ \mathbf{0.148 \pm 0.028}$   \\ \hline
boston   & $ 1.010 \pm 0.118 $           & $ \mathbf{0.931 \pm 0.107}$   \\ \hline
kin8nm   & $ 0.776 \pm 0.017 $           & $ \mathbf{0.754 \pm 0.023}$   \\ \hline
\end{tabular}
\caption{\textbf{UCI Interval Score with Recalibration.} Recalibration tends to improve the interval score.}
\label{table:uci_recal_int}
\end{center}
\end{figure*}

\begin{figure*}[h!]
\begin{center} 
\begin{tabular}{ |c|c|c|c|c|c|c| } 
 \hline
          & \textbf{\textit{Interval}}                   & \textbf{\textit{Interval Recalibrated}}      \\ \hline
concrete & $ \mathbf{0.061 \pm 0.008 }$  & $ 0.068 \pm 0.015 $  \\ \hline
power    & $ \mathbf{0.023 \pm 0.003 }$  & $ 0.028 \pm 0.008 $  \\ \hline
wine     & $ \mathbf{0.079 \pm 0.014 }$  & $ \mathbf{0.079 \pm 0.019 }$  \\ \hline
yacht    & $ \mathbf{0.121 \pm 0.005 }$  & $ 0.136 \pm 0.025 $  \\ \hline
naval    & $ \mathbf{0.043 \pm 0.014 }$  & $ 0.105 \pm 0.016 $  \\ \hline
energy   & $ \mathbf{0.060 \pm 0.010 }$  & $ 0.066 \pm 0.005 $  \\ \hline
boston   & $ 0.079 \pm 0.015 $           & $ \mathbf{0.078 \pm 0.012 }$  \\ \hline
kin8nm   & $ \mathbf{0.048 \pm 0.006 }$  & $ 0.061 \pm 0.010 $  \\ \hline
\end{tabular}
\caption{\textbf{UCI Centered Interval Calibration with Recalibration.} Recalibration tends to worsen centered interval calibration.}
\label{table:uci_recal_int_cali}
\end{center}
\end{figure*}

%% file: app_ablation.tex
\section{Ablation Study}
\subsection{Ablation Study Details}
 \label{app:ablation_study_exp_setting}

The ablation study from Section~\ref{sec:ablation} (with full results in Appendix~\ref{app:full_ablation_results}) 
investigates the effect of group batching on two methods:
\textit{Cali} (combined calibration loss, one of our proposed methods)
and \textit{SQR} (a baseline method).

For each method, we applied group batching by tuning the group batching frequency hyperparameter 
with cross-validation according to the details in Appendix~\ref{app:uci_exp_details}.

When we did not apply group batching, each batch was a uniform draw from the training dataset,
which is the default setting in most batch optimization procedures.

\subsection{Full Ablation Study Experiment Results} \label{app:full_ablation_results}
The full set of results from the ablation study presented in Section~\ref{sec:ablation}
is provided here. 
To re-iterate the purpose of this study: we show how group batching affects
UQ performance on two methods: \textit{Cali} (combined calibration loss, which is one 
of our proposed methods) and \textit{SQR} (a baseline method).

We present the effect of group batching via all of the evaluation metrics
(average calibration, sharpness, adversarial group calibration, check score, interval score,
and centered interval calibration).

\begin{figure*}[h!]
\begin{center} 
\begin{tabular}{ |c||c|c|}
 \hline & \multicolumn{2}{c|}{{\textit{Cali}}} \\
 \hline
          & \textit{Random Batch}                 & \textit{Group Batch}               \\\hline
Concrete & $6.6\pm0.9(17.6\pm2.3)$ & $\textbf{5.6}\pm0.8(\underline{17.3}\pm1.5)$ \\ \hline
Power    & $\textbf{1.7}\pm0.2(14.2\pm0.3)$ & $2.0\pm0.1(\underline{13.1}\pm0.1)$    \\ \hline
Wine     & $4.4\pm0.5(\underline{25.6}\pm0.8)$ & $\textbf{4.2}\pm0.4(26.0\pm0.8)$ \\ \hline
Yacht    & $11.1\pm1.8(\underline{1.8}\pm0.1)$ & $\textbf{8.3}\pm0.6(2.0\pm0.4)$     \\ \hline
Naval    & $2.8\pm0.2(\underline{12.1}\pm3.1)$ & $\textbf{2.4}\pm0.3(50.6\pm8.6)$    \\ \hline
Energy   & $9.2\pm0.3(\underline{2.8}\pm0.1)$ & $\textbf{5.8}\pm0.4(3.6\pm0.3)$  \\ \hline
Boston   & $9.7\pm1.3(\underline{10.2}\pm0.7)$ & $\textbf{8.5}\pm1.5(10.9\pm0.6)$ \\ \hline
Kin8nm   & $\textbf{3.4}\pm0.3(\underline{13.7}\pm0.4)$ & $3.5\pm0.3(\underline{13.7}\pm0.7)$ \\ \hline
\end{tabular}
\end{center}

\begin{center}
\begin{tabular}{ |c||c|c|}
 \hline & \multicolumn{2}{c|}{{\textit{SQR}}} \\
 \hline
          & \textit{Random Batch}      & \textit{Group Batch}   \\\hline
Concrete  & $9.8\pm1.3(\underline{7.0}\pm0.6)$ & $\textbf{7.1}\pm0.9(8.5\pm0.6)$ \\ \hline
Power     & $\textbf{2.5}\pm0.3(14.0\pm0.5)$ & $2.9\pm0.5(\underline{13.6}\pm0.8)$ \\ \hline
Wine      & $4.5\pm0.4(29.7\pm0.5)$ & $\textbf{4.0}\pm0.4(\underline{28.5}\pm0.8)$ \\ \hline
Yacht     & $9.0\pm0.9(\underline{0.9}\pm0.1)$ & $\textbf{8.9}\pm0.9(2.3\pm0.2)$ \\ \hline
Naval     & $8.6\pm1.6(\underline{3.6}\pm0.1)$ & $\textbf{5.3}\pm0.8(6.0\pm0.5)$ \\ \hline
Energy    & $10.2\pm0.8(\underline{1.8}\pm0.1)$ & $\textbf{6.9}\pm1.1(2.4\pm0.2)$ \\ \hline
Boston    & $10.9\pm1.0(\underline{8.8}\pm1.1)$ & $\textbf{9.8}\pm1.2(9.5\pm0.9)$ \\ \hline
Kin8nm    & $4.7\pm0.3(\underline{11.1}\pm0.1)$ & $\textbf{3.9}\pm0.4(11.3\pm0.2)$ \\ \hline
\end{tabular}
\end{center}
\caption{\textbf{Group Batching Ablation: Average Calibration and Sharpness.}
    The table shows mean ECE and sharpness (in parentheses) and their standard error with and without group batching. The best mean ECE for each dataset has been bolded and the best mean sharpness has been underlined for \textit{Cali} and \textit{SQR} separately. All values have been multiplied by 100 for readability.
    This is the same table as Figure~\ref{fig:group_batch_ablation_main} from Section~\ref{sec:ablation}, which is repeated here for completeness.
    \label{table:group_batch_ablation_app}}
\end{figure*}

\begin{figure*}[h!]
    \centering
    \includegraphics[width=\textwidth]{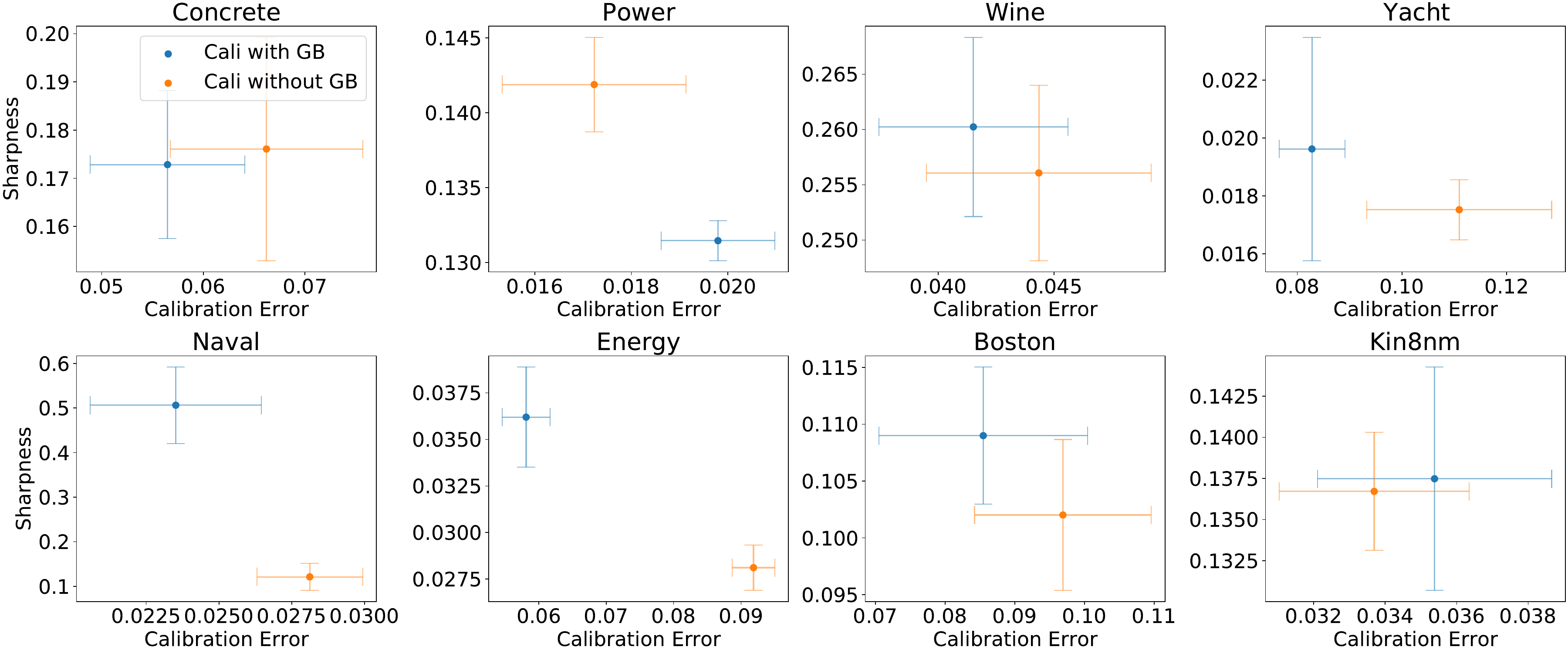}
    \caption{\textbf{Group Batching Ablation with \textit{Cali} Method: Average Calibration and Sharpness Plot.} 
    GB in legend refers to group batching. 
    Group batching tends to improve calibration and worsen sharpness for the \textit{Cali} method.}
    \label{fig:cali_ablation_cal_sharp}
\end{figure*}

\begin{figure*}[h!]
    \centering
    \includegraphics[width=\textwidth]{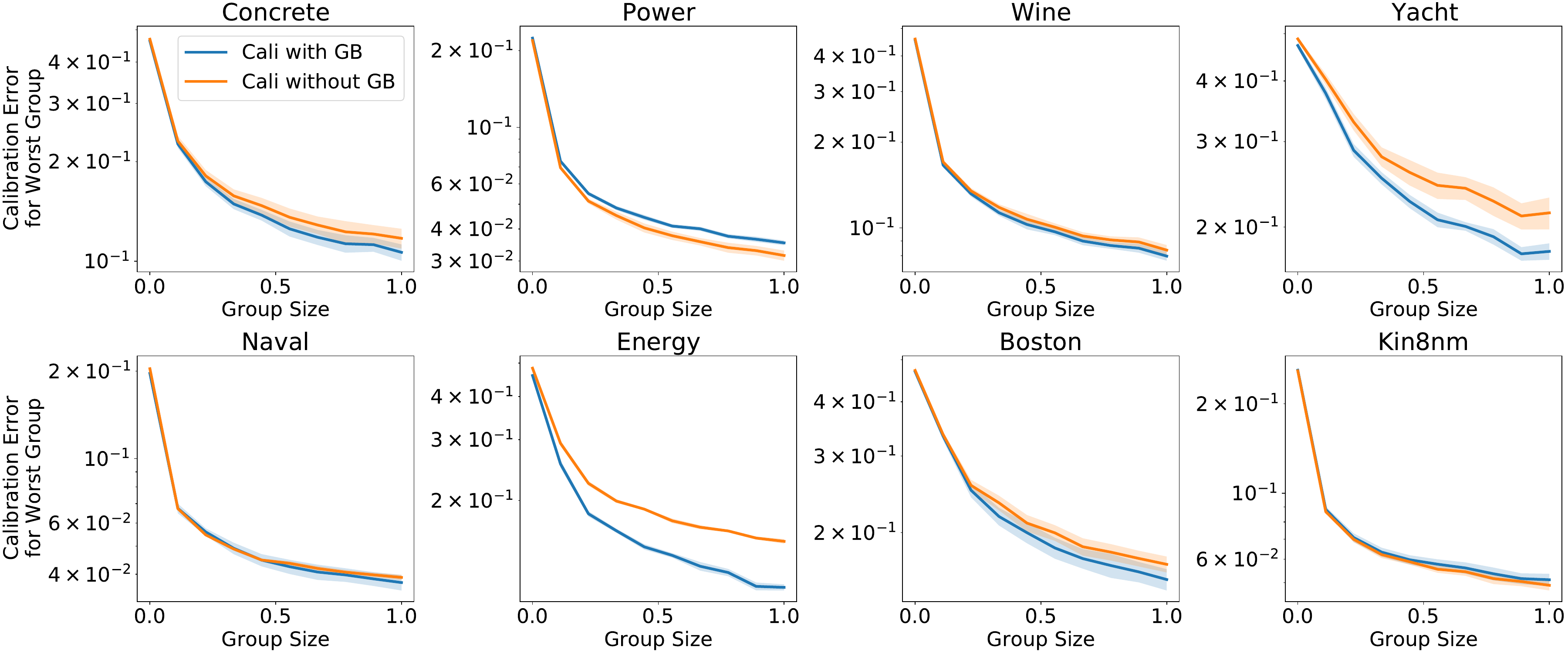}
    \caption{\textbf{Group Batching Ablation with \textit{Cali} Method: Adversarial Group Calibration.} 
    GB in legend refers to group batching. 
    Group batching tends to improve adversarial group calibration for the \textit{Cali} method.}
\end{figure*}

\begin{figure*}[h!]
    \centering
    \includegraphics[width=\textwidth]{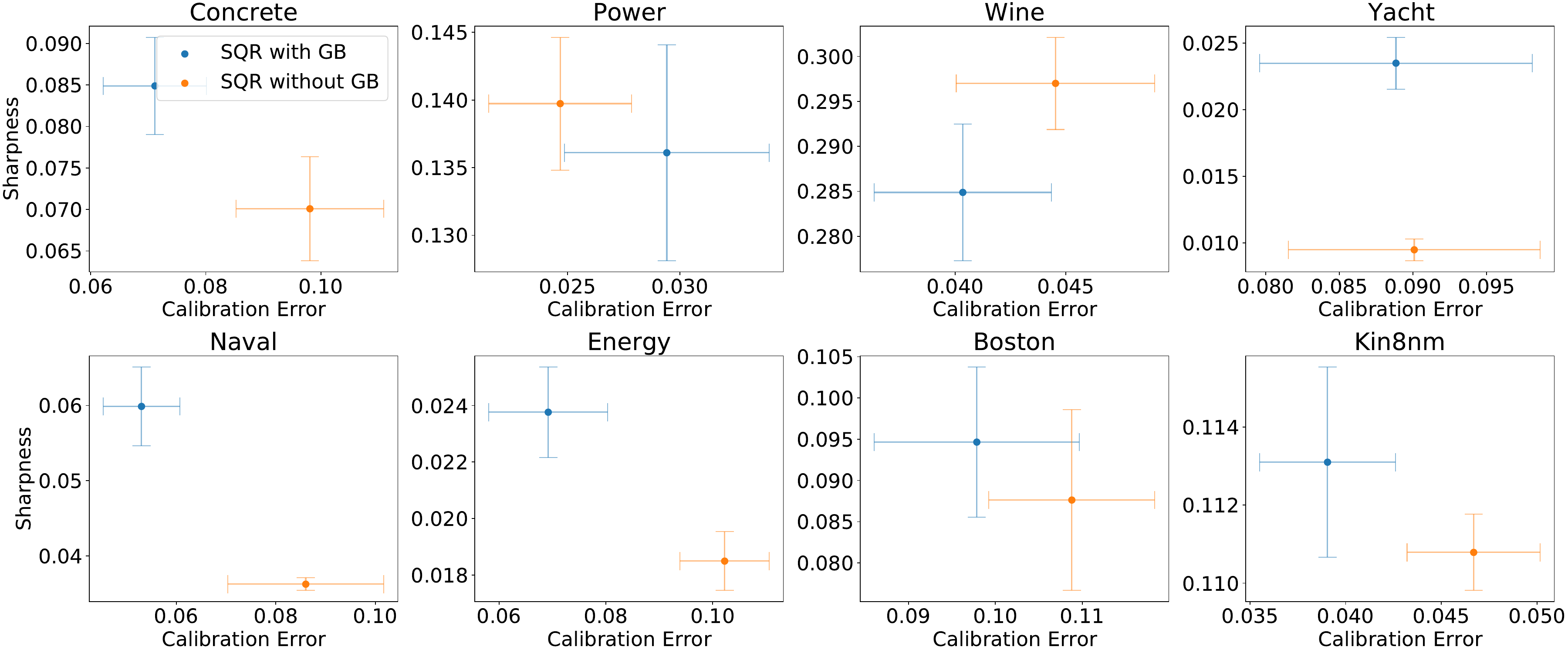}
    \caption{\textbf{Group Batching Ablation with \textit{SQR} Method: Average Calibration and Sharpness Plot.} 
    GB in legend refers to group batching. 
    Group batching tends to improve calibration and worsen sharpness for \textit{SQR} method.}
\end{figure*}

\begin{figure*}[h!]
    \centering
    \includegraphics[width=\textwidth]{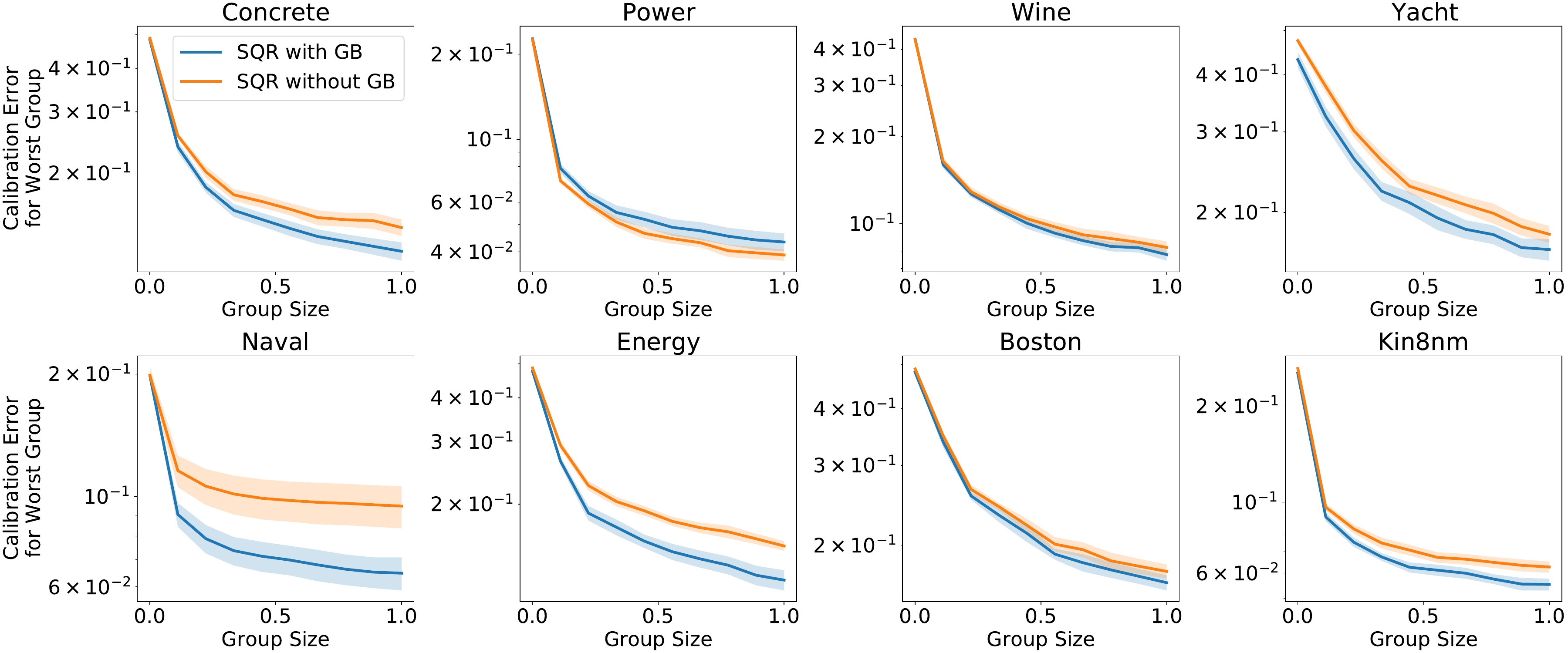}
    \caption{\textbf{Group Batching Ablation with \textit{SQR} Method: Adversarial Group Calibration.}
    GB in legend refers to group batching. 
    Group batching tends to improve adversarial group calibration for \textit{SQR} method.}
\end{figure*}

\begin{figure*}[h!]
\begin{center} 
\begin{tabular}{ |c||c|c || c|c|c|}
 \hline & \multicolumn{2}{c||}{\textit{Cali}} & \multicolumn{2}{c|}{\textit{SQR}} \\
 \hline
          & \textit{Random Batch}                 & \textit{Group Batch}            & \textit{Random Batch}      & \textit{Group Batch}   \\\hline
Concrete  & $ 0.120 \pm 0.007 $ & $ \mathbf{0.118 \pm 0.006} $   &   $ 0.083 \pm 0.006 $ & $ \mathbf{0.077 \pm 0.004} $ \\ \hline
Power     & $ \mathbf{0.063 \pm 0.002} $ & $ 0.064 \pm 0.001 $   &   $ \mathbf{0.056 \pm 0.001} $ & $ 0.057 \pm 0.001 $ \\ \hline
Wine      & $ \mathbf{0.209 \pm 0.007} $ & $ 0.210 \pm 0.008 $   &   $ \mathbf{0.206 \pm 0.008} $ & $ \mathbf{0.206 \pm 0.008} $ \\ \hline
Yacht     & $ \mathbf{0.018 \pm 0.002} $ & $ 0.019 \pm 0.004 $   &   $ \mathbf{0.011 \pm 0.002} $ & $ 0.013 \pm 0.002 $ \\ \hline
Naval     & $ \mathbf{0.042 \pm 0.009} $ & $ 0.159 \pm 0.029 $   &   $ \mathbf{0.007 \pm 0.000} $ & $ 0.014 \pm 0.001 $ \\ \hline
Energy    & $ \mathbf{0.016 \pm 0.000} $ & $ 0.017 \pm 0.002 $   &   $ \mathbf{0.013 \pm 0.000} $ & $ \mathbf{0.013 \pm 0.001} $ \\ \hline
Boston    & $ \mathbf{0.100 \pm 0.013} $ & $ 0.103 \pm 0.013 $   &   $ 0.091 \pm 0.007 $ & $ \mathbf{0.089 \pm 0.008} $ \\ \hline
Kin8nm    & $ \mathbf{0.094 \pm 0.002} $ & $ 0.096 \pm 0.005 $   &   $ 0.079 \pm 0.001 $ & $ \mathbf{0.077 \pm 0.000} $ \\ \hline
\end{tabular}
\caption{\textbf{Group Batching Ablation: Check Score.}
    This table shows mean test check score and their standard error with and without group batching
    for both \textit{Cali} and \textit{SQR}. The best mean for each dataset has been bolded for \textit{Cali} and \textit{SQR} separately. 
    While the general pattern is that group batching worsens the check score, 
    this is expected because group batching tends to worsen sharpness
    and the check score favors sharpness (Proposition 1 in main paper).
    Still, the change in the mean of the check score tends to be insignificant
    when considering the standard errors (except for Naval dataset).
    }\label{tab:group_batch_ablation_check}
\end{center}
\vspace{-5mm}
\end{figure*}

\begin{figure*}[h!]
\begin{center} 
\begin{tabular}{ |c||c|c || c|c|c|}
 \hline & \multicolumn{2}{c||}{\textit{Cali}} & \multicolumn{2}{c|}{\textit{SQR}} \\
 \hline
          & \textit{Random Batch}                 & \textit{Group Batch}            & \textit{Random Batch}      & \textit{Group Batch}   \\\hline
Concrete & $ 1.498 \pm 0.083 $  &  $ \mathbf{1.465 \pm 0.086} $   &   $ 1.254 \pm 0.120 $  &  $ \mathbf{1.079 \pm 0.066} $ \\ \hline
Power    & $ \mathbf{0.667 \pm 0.025} $  &  $ 0.699 \pm 0.019 $   &   $ \mathbf{0.603 \pm 0.016} $  &  $ 0.615 \pm 0.014 $ \\ \hline
Wine     & $ \mathbf{2.495 \pm 0.130} $  &  $ 2.498 \pm 0.135 $   &   $ \mathbf{2.325 \pm 0.107} $  &  $ 2.362 \pm 0.117 $ \\ \hline
Yacht    & $ \mathbf{0.276 \pm 0.040} $  &  $ 0.298 \pm 0.063 $   &   $ 0.177 \pm 0.033 $  &  $ \mathbf{0.164 \pm 0.029} $ \\ \hline
Naval    & $ \mathbf{0.479 \pm 0.098} $  &  $ 1.560 \pm 0.268 $   &   $ \mathbf{0.069 \pm 0.003} $  &  $ 0.144 \pm 0.014 $ \\ \hline
Energy   & $ 0.218 \pm 0.009 $  &  $ \mathbf{0.204 \pm 0.018} $   &   $ 0.191 \pm 0.006 $  &  $ \mathbf{0.172 \pm 0.012} $ \\ \hline
Boston   & $ \mathbf{1.437 \pm 0.255} $  &  $ 1.449 \pm 0.259 $   &   $ 1.284 \pm 0.151 $  &  $ \mathbf{1.217 \pm 0.152} $ \\ \hline
Kin8nm   & $ \mathbf{1.102 \pm 0.031} $  &  $ 1.121 \pm 0.072 $   &   $ 0.914 \pm 0.016 $  &  $ \mathbf{0.871 \pm 0.011} $ \\ \hline
\end{tabular}
\caption{\textbf{Group Batching Ablation: Interval Score.}
    This table shows mean test interval score and their standard error with and without group batching
    for both \textit{Cali} and \textit{SQR}. The best mean for each dataset has been bolded for \textit{Cali} and \textit{SQR} separately. 
    The general pattern is that the interval score worsens for \textit{Cali} 
    and improves for \textit{SQR}.
    However, the change tends to be insignificant
    when considering the standard errors (except for Naval dataset).}
\end{center}
\vspace{-5mm}
\end{figure*}

\begin{figure*}[h!]
\begin{center} 
\begin{tabular}{ |c||c|c || c|c|c|}
 \hline & \multicolumn{2}{c||}{\textit{Cali}} & \multicolumn{2}{c|}{\textit{SQR}} \\
 \hline
          & \textit{Random Batch}                 & \textit{Group Batch}            & \textit{Random Batch}      & \textit{Group Batch}   \\\hline
Concrete    & $ 0.102 \pm 0.014 $ & $ \mathbf{0.096 \pm 0.013} $   &   $ 0.188 \pm 0.029 $ & $ \mathbf{0.127 \pm 0.014} $ \\ \hline
Power       & $ \mathbf{0.033 \pm 0.003} $ & $ 0.037 \pm 0.002 $   &   $ \mathbf{0.040 \pm 0.008} $ & $ 0.047 \pm 0.006 $ \\ \hline
Wine        & $ 0.072 \pm 0.007 $ & $ \mathbf{0.065 \pm 0.007} $   &   $ 0.057 \pm 0.005 $ & $ \mathbf{0.055 \pm 0.005} $ \\ \hline
Yacht       & $ 0.139 \pm 0.018 $ & $ \mathbf{0.129 \pm 0.016} $   &   $ 0.128 \pm 0.027 $ & $ \mathbf{0.119 \pm 0.020} $ \\ \hline
Naval       & $ 0.048 \pm 0.006 $ & $ \mathbf{0.034 \pm 0.002} $   &   $ 0.114 \pm 0.032 $ & $ \mathbf{0.066 \pm 0.009} $ \\ \hline
Energy      & $ 0.146 \pm 0.013 $ & $ \mathbf{0.090 \pm 0.011} $   &   $ 0.171 \pm 0.018 $ & $ \mathbf{0.104 \pm 0.020} $ \\ \hline
Boston      & $ 0.142 \pm 0.032 $ & $ \mathbf{0.138 \pm 0.028} $   &   $ 0.195 \pm 0.021 $ & $ \mathbf{0.173 \pm 0.027} $ \\ \hline
Kin8nm      & $ \mathbf{0.061 \pm 0.004} $ & $ 0.067 \pm 0.005 $   &   $ 0.079 \pm 0.003 $ & $ \mathbf{0.069 \pm 0.009} $ \\ \hline

\end{tabular}
\caption{\textbf{Group Batching Ablation: Centered Interval Calibration.}
    This table shows mean test centered interval calibration (measured by ECE for centered intervals) score and their standard error with and without group batching
    for both \textit{Cali} and \textit{SQR}. The best mean for each dataset has been bolded for \textit{Cali} and \textit{SQR} separately. 
    The general pattern is that group batching improves centered interval calibration 
    for both \textit{Cali} and \textit{SQR}.
    While the change tends to be insignificant
    when considering the standard errors,
    the improvement is significant in numerous cases 
    (e.g. Naval and Energy for both methods, Concrete with \textit{SQR}).}
\end{center}
\end{figure*}

%% file: app_epistemic.tex
\section{Considerations for Epistemic Uncertainty} \label{app:epistemic}

\subsection{Sources for Epistemic Uncertainty}
The primary focus of this paper is on learning a quantile model.
For any single setting of the parameters of the quantile model, the model outputs the 
current best estimate of the \textit{true underlying distribution} of the dataset.
Following the notation laid out in Section~\ref{sec:prelim}, the learned quantile model $\hQ$
is a best approximation to $\Q$, the quantile function of the true distribution.

Meanwhile, epistemic uncertainty refers to the uncertainty \textit{in making the distributional prediction, $\hQ$}.
\citet{pearce2018high} provides one method of decomposing the sources of epistemic uncertainty in a regression setting:
\begin{itemize}
        \item \emph{Model misspecification:} $\hQ$ may lack the flexibility to
        accurately model $
        \Q$, leading to systematic bias.
        
        \item  \emph{Data uncertainty:} $\hQ$ may not be estimated using a
        representative sample $\{ x_i, y_i \}$ from the assumed underlying distribution.
        
        \item \emph{Parameter uncertainty:} $\hQ$ may not be estimated using
        a large enough quantity of samples, leading to uncertainty about the estimated quantity.
\end{itemize}

\citet{pearce2018high} has argued that, 
given the rich class of function approximators 
at hand today (NN, deep trees, ensembles), model misspecification can be ignored, which we believe is reasonable. 
In modeling the remaining sources of uncertainties in $\hQ$, 
we can incorporate common standard methods to quantify epistemic uncertainty, 
including boostrapping the data, creating an ensemble of estimates for $\hQ$ with random parameter initializations, or fitting a residual process \citep{liu2019accurate}.  
Here, we describe one combination of these methods: 
an ensemble of estimates of the learned quantile function 
$\{\hQ^{(1)}, \hQ^{(2)},...\}$ each trained with random initialization (to address parameter uncertainty), on a bootstrapped sample of the training data 
(to address data uncertainty).
The uncertainty over this set of models is the epistemic uncertainty.

\subsection{Expressing and Utilizing Epistemic Uncertainty} \label{app:boot_ens}
Once we decide on methods to quantify the epistemic uncertainty, 
the next question is \textit{how to express the epistemic uncertainty}, 
especially when combining it with the current prediction of the aleatoric uncertainty.
This is still an open research question, especially in the regression setting, 
and methods of combining aleatoric with epistemic uncertainty will differ 
for how the uncertainty is represented (e.g. density estimates, quantiles, prediction intervals). 

One method of combination is to consider the utility of quantifying epistemic uncertainty.
Intuitively, for a given input, if we have high epistemic uncertainty, 
the combined uncertainty should be higher (i.e. less confident prediction), and vice versa.
If we consider a single quantile, it is unclear whether lower confidence
(due to epistemic uncertainty) would equate to a lower or higher quantile estimate.
However, if we consider constructing a centered prediction interval for total uncertainty,
it is straightforward to see that lower confidence should widen the interval, 
by raising the upper bound (quantile level above $0.5$) 
and lowering the lower bound (quantile level below $0.5$).
This conservative upper and lower bound can be constructed with 
the bootstrap distribution of each quantile according to each ensemble member prediction.
This is also the method utilized in \citet{pearce2018high}.

Suppose we have an ensemble of $M$ quantile models: 
$\{\hQ^{(1)}, \hQ^{(2)}, \dots, \hQ^{(M)}\}$.
For any test point $x^{*}$ and test coverage level $(1-\alpha^{*})$,
the total uncertainty represented by a centered prediction interval with upper bound $\hat{u}$ and lower bound $\hat{l}$ is constructed as:
\begin{align*}
    \hat{l} &= \bar{\mu}(x^{*}, \alpha^{*}/2) - z \frac{s(x^{*}, \alpha^{*}/2)}{\sqrt{M}} \\
    \hat{u} &= \bar{\mu}(x^{*}, 1 - \alpha^{*}/2) + z \frac{s(x^{*}, 1 - \alpha^{*}/2)}{\sqrt{M}} \\
    &\bar{\mu}(x^{*}, p^*) = \frac{1}{M}\sum_{i=1}^{M} \hQ^{(i)}_{p^{*}}(x^{*}) \\
    &s(x^{*}, p^*) = \sqrt{\frac{1}{M-1}\sum_{i=1}^{M} (\hQ^{(i)}_{p^{*}}(x^{*}) - \bar{\mu}(x^{*}, p^{*}))^{2}}
\end{align*}
and $z$ is the chosen critical value (e.g. $1.96$ for a conservative bound that takes the $95\%$ confidence interval of the bootstrap distribution).
In words, the construction of $\hat{u}, \hat{l}$ equates to constructing a
conservative prediction interval that depends on how dispersed or concentrated each ensemble member's predictions are.

\subsection{Metrics for Total Uncertainty Evaluation} \label{app:epist_metric}
After choosing a method to express epistemic uncertainty and combining it with 
aleatoric uncertainty for a prediction of total uncertainty, 
next comes the question of \textit{how to evaluate the combined uncertainty}.

The critical point here is that the calibration, sharpness and 
proper scoring rule metrics we have discussed thus far \textit{are not applicable here}. 
This is because these metric only judge how close the prediction is to the true
underlying distribution.
In fact, ECE (a measure of average calibration) can be shown to be identical to
the Wasserstein distance between distributions under the $L1$ distance metric \citep{zhao2020individual}.
In a hypothetical setting where we have very few datapoints and hence 
very high epistemic uncertainty throughout the whole data support, 
if one distributional prediction was extremely lucky and predicted a distribution that
adheres exactly to the true underlying distribution, the aforementioned metrics
will consider this prediction a perfect prediction -- 
however, a lucky guess is not at all a useful UQ, and to a practitioner,
a less confident prediction (by quantifying high epistemic uncertainty) is much more useful, 
rather than a very confident prediction that can be correct if it is lucky, 
but confidently very incorrect otherwise.

We also emphasize here that, while standard evaluation experiments and metrics exist for the
\textit{classification} setting (e.g. by training an image classifier on the MNIST dataset
and testing on the Non-MNIST dataset to assess the entropy of the predicted class probabilities or 
the output of a trained out-of-distribution detector), 
there does not exists standard experiments and metrics to evaluate epistemic uncertainty in the  \textit{regression setting}.

Therefore, we propose one evaluation metric to assess combined total uncertainty, which is \textit{sharpness subject to sufficient coverage} and we will refer to this metric as ``epistemic coverage''.
Epistemic coverage measures average calibration of a centered prediction interval, but the difference in observed probabilities from expected probabilities is penalized only when the observed probability is less than
the expected probability (i.e. do not penalize the prediction if the observed probability is
higher than the expected probability), and if this sufficient coverage condition is 
met, then we evaluate sharpness. 

By this metric, only over-confidence is penalized and under-confidence 
is considered acceptable. However, since infinitely wide prediction intervals 
are also not useful, we also consider sharpness after sufficient coverage is met.

\subsection{Demonstrating Effect of Bootstrap Ensembling and the Epistemic Coverage Metric}
We design an ``epistemic experiment'' to show the effect of incorporating epistemic uncertainty via 
bootstrapped ensembles.
In this experiment, we swap the train and test sets, such that the training set is much smaller than the test set (roughly $\frac{1}{7}$ of test set size), hence, the model should have very high epistemic uncertainty in making predictions on the test set.
It is expected that by not incorporating epistemic uncertainty in such a setting, 
the model will produce \textit{overconfident} predictions that are \textit{too sharp}.
This overconfidence will be penalized heavily by the epistemic coverage metric we described above in Appendix~\ref{app:epist_metric}, and sharpness should also indicate that the predictions are 
too tight.
Producing conservative distributional predictions via the bootstrapped ensembling technique described above 
in Appendix~\ref{app:boot_ens} is expected to mitigate these overconfidence issues
by incorporating epistemic uncertainty.

\begin{figure*}[h!]
    \centering
    \includegraphics[width=\textwidth]{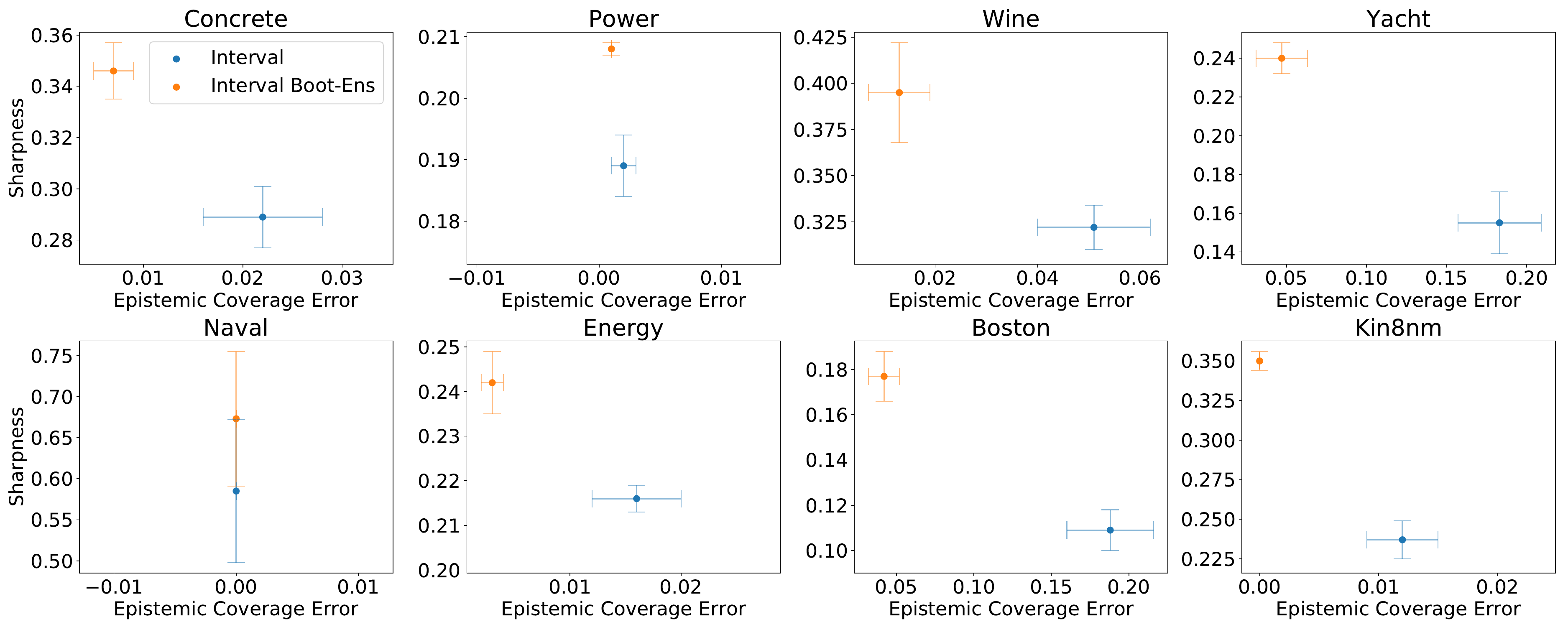}
    \caption{\textbf{Epistemic Experiments on UCI Datasets.} We evaluate epistemic coverage in
    the epistemic experiment setting where the train set is much smaller than the test set. Epistemic coverage only penalizes overconfidence. Sharpness is the average width of the 95\% PI.}
    \label{fig:epist_uci_cal_sharp}
\end{figure*}

We show the effect on one of our methods, \textit{Interval}, because the same effect can be observed for any of the quantile methods, including the baseline algorithms.
The results are shown in Figure~\ref{fig:epist_uci_cal_sharp}.

When a conservative PI is constructed with the bootstrapped ensemble (labelled \textit{Interval Boot-Ens}), the epistemic coverage error decreases significantly to or near zero, which is expected given that the conservative bounds only work to widen the PI, and the epistemic coverage error only penalizes PI that are not wide enough.
The increase in width is also evident in the increase in sharpness with bootstrap ensembling.

Therefore, the ensembling technique does work to incorporate epistemic uncertainty
from a practical standpoint by imbuing more underconfidence into the distributional predictions.
Still, quantifying and evaluating epistemic uncertainty in a regression setting is an 
open problem, and we leave for future work developing alternative methods of quantifying 
epistemic uncertainty in regression.

%% file: app_add_discussion.tex
\section{Additional Discussions}

\paragraph{Potential negative social impacts}

This work proposes methods in uncertainty quantification (UQ), 
with a focus on the notion of calibration. 
UQ is a field that is becoming more and more important as many autonomous systems are being deployed in various real-life applications (self-driving cars, security devices, object recognition systems).
While we believe the development of robust and
accurate methods in UQ will accelerate deployment of autonomous systems and expand real-world 
use-cases, 
we also acknowledge the potential disruptive effects 
such change can have in the relevant industries.

Futher, calibration is a relevant notion in fairness \citep{kleinberg2016inherent}.
Though we propose methods to achieve better calibration
when making probabilistic forecasts, 
we note the potential for using such information with 
ill intent, e.g. to intentionally avoid calibrated (or fair) 
decisions.

\paragraph{Assets used in this work}
In implementing our work, we have referenced the implementation of 
one of the baseline methods (\textit{SQR}), which is publicly available under the Creative Commons Attribution-NonCommercial 4.0 International Public License.

We also state that no data from human subjects were used in this work and 
thus there is no personal identifying information.